\documentclass[10pt,twocolumn,letterpaper]{article}

\usepackage{iccv}
\usepackage{times}
\usepackage{epsfig}
\usepackage{graphicx}
\usepackage{amsmath}
\usepackage{amssymb}
\usepackage{booktabs}
\usepackage{multirow}
\usepackage{subfigure}
\usepackage{bbding}
\usepackage{booktabs}
\usepackage{makecell}

\def \ie{{\em i.e.}}
\def \eg{{\em e.g.}}

\usepackage[pagebackref=true,breaklinks=true,letterpaper=true,colorlinks,bookmarks=false]{hyperref}

\iccvfinalcopy % *** Uncomment this line for the final submission

\def\iccvPaperID{3351} % *** Enter the ICCV Paper ID here

\begin{document}
	
	%%%%%%%%% TITLE
	\title{Improving Contrastive Learning by Visualizing Feature Transformation}

	%{\tt\small zhurui.kimsoybean@gmail.com}

	\author{Rui Zhu$^{1,2}$\thanks{Equally-contributed and this work is done at JD AI Research.},~~ Bingchen Zhao$^{3*}$,~~ Jingen Liu $^2$\thanks{Corresponding author.},~~ Zhenglong Sun$^1$,~~ Chang Wen Chen$^4$  \\
		{\small\centering$^1$ The Chinese University of HongKong, Shenzhen}~~~
		{\small\centering$^2$ JD AI Research}~~~
		{\small\centering$^3$  Tongji University}~~~
		{\small\centering$^4$  The Hong Kong Polytechnic University}
		\\
		{\tt\scriptsize ruizhu@link.cuhk.edu.cn, \{zhaobc.gm, jingenliu\}@gmail.com, sunzhenglong@cuhk.edu.cn, changwen.chen@polyu.edu.hk}
	}
	
	\maketitle
	%\thispagestyle{empty}
	
	%%%%%%%%% ABSTRACT
	\begin{abstract}
		Contrastive learning, which aims at minimizing the distance between positive pairs while maximizing that of negative ones, has been widely and successfully applied in unsupervised feature learning, where the design of positive and negative (pos/neg) pairs is one of its keys. In this paper, we attempt to devise a feature-level data manipulation, differing from data augmentation, to enhance the generic contrastive self-supervised learning. To this end, we first design a visualization scheme for pos/neg score\footnote{Pos/neg score indicates cosine similarity of pos/neg pair.} distribution, which enables us to analyze, interpret and understand the learning process. To our knowledge, this is the first attempt of its kind. More importantly, leveraging this tool, we gain some significant observations, which inspire our novel Feature Transformation proposals including the extrapolation of positives. This operation creates harder positives to boost the learning because hard positives enable the model to be more view-invariant. Besides, we propose the interpolation among negatives, which provides diversified negatives and makes the model more discriminative. 
		It is the first attempt to deal with both challenges simultaneously. Experiment results show that our proposed Feature Transformation can improve at least $6.0\%$ accuracy on ImageNet-100 over MoCo baseline, and about $2.0\%$ accuracy on ImageNet-1K over the MoCoV2 baseline. Transferring to the downstream tasks successfully demonstrate our model is less task-bias. Visualization tools and codes: \url{https://github.com/DTennant/CL-Visualizing-Feature-Transformation}.
	\end{abstract}
	
	%%%%%%%%% BODY TEXT
	\section{Introduction}
	
	Finetuning from ImageNet \cite{imagenet} supervised pre-train networks \cite{res,vgg,densely} for down-stream tasks, such as object detection \cite{ssd,YOLo,faster} and semantic segmentation \cite{long2015fully,chen2014semantic}, is a de facto dominant approach in computer vision community. But recently self-supervised contrastive learning achieves comparable transfer performance without the human-provided annotations. One of the key issues of contrastive learning is to design positive and negative (pos/neg) pairs to learn an embedding space such that the positives stay closer in the space while the negatives are pushed away.  
	
	\begin{figure}
		\centering
		\subfigure[Observation]{
			\label{fig:observation}
			\includegraphics[width=0.31\linewidth,height=0.31\linewidth]{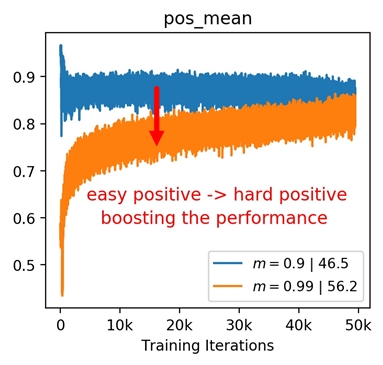}}
		\subfigure[Proposed Method]{
			\label{fig:proposed_method}
			\includegraphics[width=0.31\linewidth,height=0.29\linewidth]{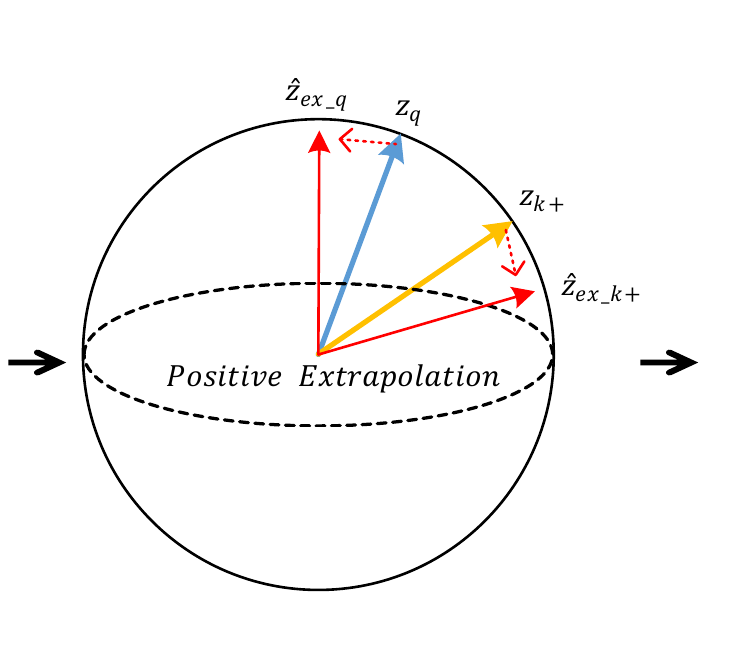}}
		\subfigure[Performance Gain]{
			\label{fig:performance_gain}
			\includegraphics[width=0.31\linewidth,height=0.31\linewidth]{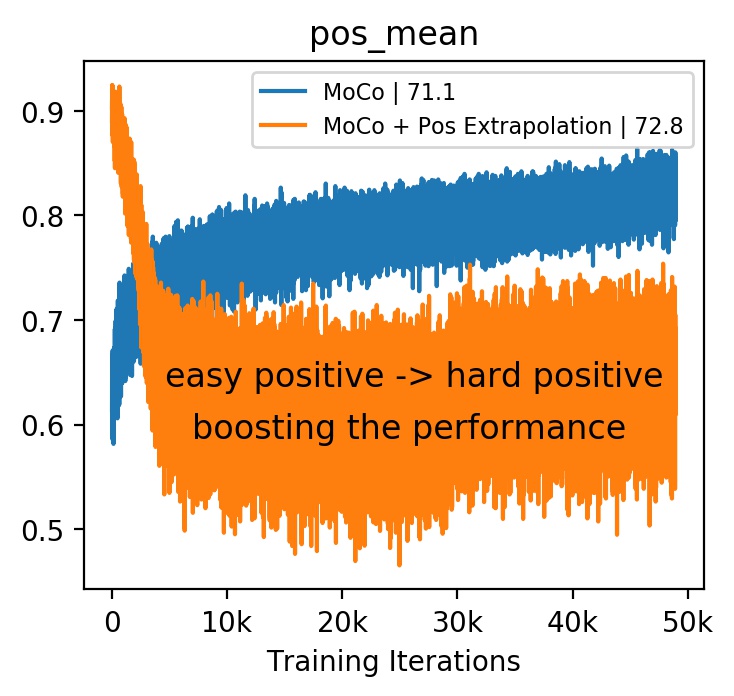}}
		\caption{The motivation of visualizing the score distribution. (a) It draws the score distribution of positive pairs for m (the momentum in MoCo\cite{he2020momentum}) being 0.99 and 0.9, showing that smaller positive scores generally need longer time to converge and obtain better accuracy. (b) Inspired by (a), we apply extrapolation on positive pairs to slightly decrease the scores, generating harder positives. (c) Leveraging the extrapolation of positives, we improve the performance from 71.1\% (the blue) to 72.8\% (the orange). The performance increase is consistent with the change of distribution. The mean score of positive pairs changes from blue plot (before extrapolation) to orange plot (after extrapolation).}
		\label{fig:intro}
		\vspace{-0.4cm}
	\end{figure}

	Most existing approaches \cite{caron2020unsupervised,chen2020simple,tian2020makes,chen2020improved} acquire pos/neg pairs by data augmentation, which exploits various views of the same image to form positive pairs. For example, CMC\cite{tian2019contrastive} uses the luminance and chrominance color channel of an image as two views. InfoMin \cite{tian2020makes} demonstrates that incremental data augmentations indeed lead to decreasing mutual information between views and thus improve transfer performance. In other words, an effective positive pair prefers to convey more variance of one instance. With a series of promotions, the contrastive learning methods based on data augmentations \cite{caron2020unsupervised,chen2020simple,tian2020makes,chen2020improved} are achieving closer to the fully supervised performance on ImageNet\cite{chen2020simple}.   
	
	Most previous data augmentations (\eg, cropping, color distortion) are directly sourced from human intuitions, which may lack much interpretability, thus they can not guarantee their effectiveness. We argue, however, that the feature-level data manipulation (\ie, feature transformation) can provide more explainable or effective pos/neg pairs to enhance the feature embedding. To this end, we first design a scheme to visualize the pos/neg pair score distributions during the training. We believe that, from these score distributions, we can reveal and explain how the model parameter values affect its performance. The visualization can help us trace back the training process. Moreover, it enables us to observe the characteristics of the pos/neg pairs, and then invent more effective feature transformations (\textbf{FT}).

	Figure \ref{fig:intro} demonstrates the motivation of score visualization. By plotting the score distributions under different momentum values of MoCo \cite{he2020momentum}, we can clearly observe that the case of $m=0.99$ has smaller positive scores while achieves better performance. A small positive score indicates less similarity between the pair, which means this positive pair actually carrying  large view variance of one example. Actually, this is consistent with the goal of feature learning, which targets at a more view-invariant visual representation. Therefore, we conjecture that ``hard positives'' are the ones conveying large view variance of a sample. Inspired by this observation, we introduce an extrapolation operation on positive pairs to increase view variance and thus acquire hard positives. Figure \ref{fig:performance_gain} shows that the extrapolation of positives can boost the model performance from the ``blue'' one to the ``orange'' one.

	Besides, to make full use of negative features,
	we propose the random interpolation among negatives, which intuitively provides diversified negatives for each training step and makes the model more discriminative.

	Unlike the traditional data augmentation, our feature transformation does not bring additional training examples. Instead, it aims at reshaping the feature distribution by manipulating both positive and negative pairs. Basically, our feature transformation will create  hard positives and diversified negatives to learn a more view-invariant (hard positive) and a more discriminative (diversified negatives) representation. It is directly driven by the performance of the learned representation, while data augmentation is kind of blind to the performance. 
	Furthermore, our feature transformation makes the model less ``task-bias'', which means we can achieve performance improvement for various downstream tasks. It has been verified by our experiments on object detection, instance segmentation,  and long-tailed classification with significant improvement.

	Both our visualization tool and feature transformation are generic, and can be applied to various self-supervised contrastive learning including MoCo\cite{he2020momentum}, SimCLR\cite{chen2020simple}, InfoMin\cite{tian2020makes}, SwAv\cite{caron2020unsupervised}, SimSiam\cite{chen2020exploring}. In the following sections, we employ the classic model MoCo to demonstrate our framework. To summarize, our contributions include:  	
	\begin{itemize}
		\item We are the first to design a visualization tool to analyze and interpret how the score distribution of pos/neg pairs affects the model's capability. The visualization also helps us come into some significant observations.
		\item Inspired by the observations on the model visualization, we propose a simple yet effective feature transformation, which creates both ``hard positives'' and ``diversified negatives'' to enhance the training. The feature transformations enable to learn more ``view-invariant'' and discriminative representations. 
		\item We conduct thorough experiments and our model achieves the state-of-the-art performance. In addition, the experiments on the downstream tasks successfully demonstrate our model is less task biased. 
	\end{itemize}

	\section{Related Work}
	
	{\noindent \textbf{Contrastive Learning:}}
	Contrastive losses have been widely used in self-supervised learning and brought significant improvements on classification ~\cite{gutmann2010noise,bachman2019learning,he2020momentum,tian2019contrastive,tian2020makes,chen2020simple,chen2020improved,grill2020bootstrap, caron2020unsupervised,hu2021adco,cai2020joint,zhao2020makes,wei2020co2, xiao2020should,chuang2020debiased,cao2020parametric,tian2019contrastive,wang2020unsupervised,wu2018unsupervised,ye2019unsupervised,wang2021unsupervised,wang2020understanding,koohpayegani2021mean} and detection \cite{wang2020dense,xie2021detco,xie2020propagate,yang2021instance}.
	InfoMin \cite{tian2020makes} uses the lower bound of NCE to demonstrate that incremental data augmentations lead to decreasing mutual information between views and thus improve transfer performance. In other words, relatively harder data augmentation for contrastive learning  boosts the transfer performance\cite{kalantidis2020hard_mochi,chen2020simple}.
	We show that our proposed feature transformation can be easily adopted on current state-of-the-art models.

	{\noindent \textbf{MixUp for contrastive learning}}
	Mixup~\cite{zhang2017mixup} and its numerous variants~\cite{verma2019manifoldmix,yun2019cutmix,kim2020mixco} provide highly effective data augmentation strategies when paired with a cross-entropy loss for supervised and semi-supervised learning. 
	Manifold mixup~\cite{verma2019manifoldmix} is a feature-level regularization for supervised learning while Un-mix~\cite{shen2020mix} proposes using mixup in the image/pixel space for self-supervised learning; 
	And in MoChi~\cite{kalantidis2020hard_mochi} the authors propose mixing the negative sample in the embedding space for hard negatives augmentation but hurt the classification accuracy.
	i-Mix~\cite{lee2021imix} proposed a strategy mixing instances in both input and virtual label spaces to regularize contrastive training.
	In this paper, we proposed to use feature transformation rather than data augmentation. Positive features are extrapolated to increase the hardness of positives, and negative features in the memory queue are interpolated to increase the diversity. Our FT provides more efficacy compared with augmentations.
	
	{\noindent \textbf{Generating examples for metric learning:}}
	The idea of generating new examples for metric learning has been explored by ~\cite{dvml,daml,embed_expan}. The Embedding Expansion~\cite{embed_expan} work uses uniform interpolation between two positive and negative points, creates a set of synthetic points, and then selects the hardest pair as negative.
	\cite{dvml,daml} generate new hard examples by generators and improve performance for metric learning.
	Different from the approaches \cite{dvml,daml} for supervised metric learning, our pos/neg FTs are aiming at self-supervised learning and doesn't require labels, extra parameters and loss terms to be optimized.
	
	\begin{figure}
		\begin{center}
			\vspace{-0.3cm}
			\includegraphics[width=1.0\linewidth]{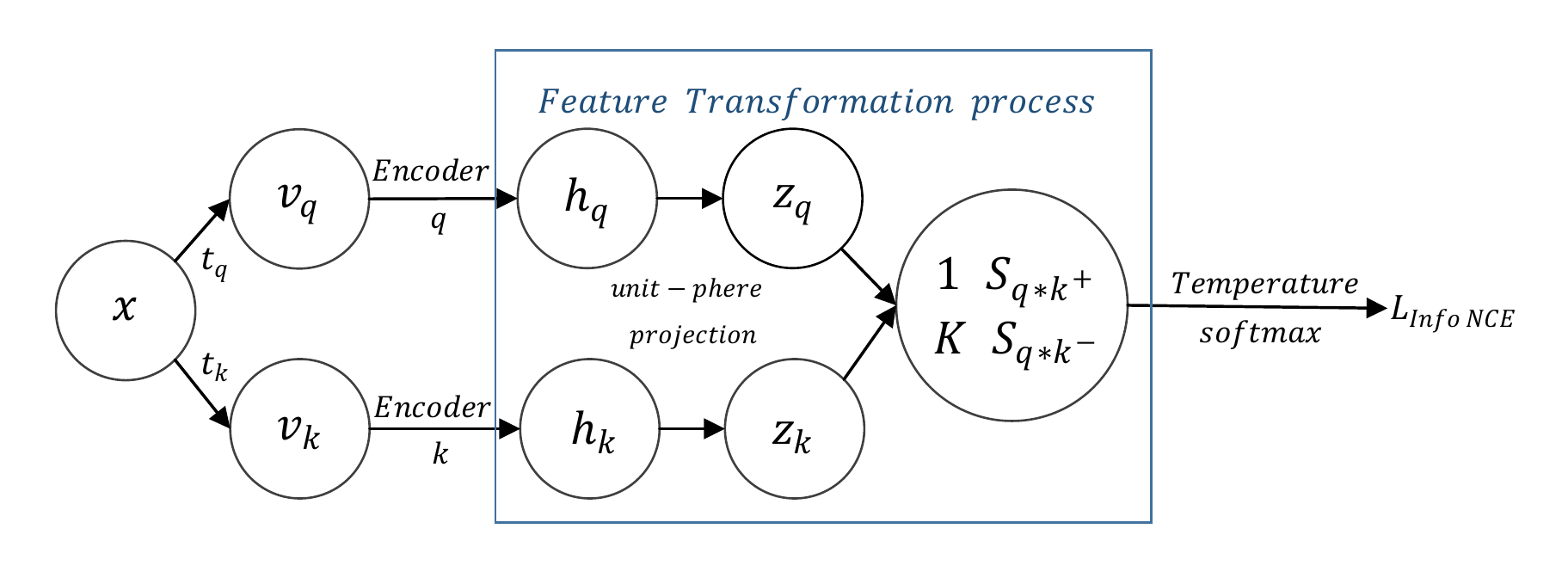}
		\end{center}
		\vspace{-0.5cm}
		\caption{Feature Transformation Contrastive learning pipeline.}
		\label{fig:pipeline}
		\vspace{-0.4cm}
	\end{figure}

	\section{Visualization of Contrastive Learning}
	\subsection{Preliminaries}\label{sec:preliminaries}
	Let us start from the basic procedures of contrastive learning, as shown in Figure \ref{fig:pipeline}. 
	Each data sample $x$ passes through two separate data augmentation pipeline $t_q$ and $t_k$, which are randomly sampled from the same data augmentation pool, and two views $v_q$ and $v_k$ will be acquired to construct positive pairs \cite{chen2020simple,grill2020bootstrap}. 
	The encoder $q$ and $k$ \footnote{Encoder $q$ and $k$ might be the same  \cite{chen2020simple} or different network \cite{he2020momentum,grill2020bootstrap}.} will respectively map two views into feature embedding space. 
	An $\ell_2$ normalization  is applied on feature vector $h_q$ and $h_{k}$ to project the corresponding vector $h_q$ and $h_{k}$ (\ie, $z_q=h_q/\|h_q\|_2$) onto the unit sphere and obtain $z_q$ and $z_{k}$. Their inner product will produce the cos similarity score, namely one positive pair score $S_{q\cdot k^{+}}$ and $K$ negative pair scores $S_{q\cdot k^{-}}$.
	These pair scores are input to InfoNCE loss \ref{eq:infonce} for contrastive learning: 
	\begin{equation} \label{eq:infonce}
	\small
	\mathcal{L}=\\- \log\left[\frac{\exp\left(S_{q\cdot k^{+}} / \tau\right)}{\exp\left(S_{q\cdot k^{+}} / \tau\right) + \sum_{K} \exp\left(S_{q\cdot k^{-}} / \tau\right)} \right] \,
	\end{equation}
	Here we roughly defined Feature Transformation process as certain manipulations on encoder embeddings $h_q$ and $h_{k}$, in order to reshape the distribution of the output pos/neg pair score ($S_{q\cdot k^{+}}$ and $S_{q\cdot k^{-}}$.), for better contrastive learning in the follow-up InfoNCE loss.  The most common FT applied in current SOTA is the  \cite{caron2020unsupervised,chen2020simple,tian2020makes,chen2020improved,he2020momentum} unit-sphere projection of $\ell_2$ normalization. We provide  empirical studies of this regular FT and illustrate it importance for significant constriction of feature length ($\ell_2$) in Supp \textbf{F}.

	\subsection{Score Distribution Visualization}\label{sec:score_vis}
	We choose to visualize the score distribution of pos/neg pairs instead of the loss curves and transfer accuracy, as the inside training dynamics can unearth the learning capability of the model.
	Specifically, there are two practical reasons:
	(1) The basic idea of InfoNCE loss is to compare the pos/neg scores in a log-softmax manner, so visualizing the input score pairs can help study the contrastive learning process. 
	(2) The normalized feature vectors $z_q$ and $z_{k}$ are high-dimensional, which is challenging for storage and visualization; The exponential amplification of scores is too large to observe the details of characteristics of pos/neg scores. However, $S_{q\cdot k}$ is one-dimensional and limited to $[-1,1]$, which is suitable to observe inside the contrastive process.

	Notice that this practical visualization tool is offline and doesn't affect training speed with negligible computation. Even with larger datasets and batch size, it’s still feasible. The details of the visualization tool are present in Supp \textbf{A}.

	\subsection{Visualization Examples with MoCo}\label{sec:visual_example}
	We choose the computationally-efficient model, MoCo \cite{he2020momentum} as an example to demonstrate our visualization design. 
	
	{\noindent \textbf{Momentum Update Mechanism:}}
	Memory queue \cite{wu2018unsupervised} is an initial approach for solving the large batch computational burden which stores $K$ negative features in the memory that will be updated using the output of the encoder at each training step. However, the rapid change of the encoder ($f_q$ and $f_k$)  could bring inconsistency into the memory queue which usually contains outdated features.
	MoCo solves the inconsistency issue by leveraging a momentum update mechanism \cite{tarvainen2017mean} where only $f_q$ is updated by back-propagation and the $f_k$ is updated by momentum mechanism:
	\begin{equation} \label{eq:momentum}
	\theta_{f_k} \leftarrow m \theta_{f_k} + (1-m) \theta_{f_q}
	\end{equation}
	where $m\in [0,1)$ is the momentum coefficient and has a huge influence to the final transfer accuracy.
	The memory queue is then updated using the features from $f_k$ because the momentum update of $f_k$ brings a smoother change of features that could reduce the inconsistency in memory queue.

	\begin{table}
		\vspace{-.5em}
		\centering
		\small
		\setlength\tabcolsep{2pt} % default value: 6pt
		\begin{tabular}{c|cccccccc}
			\toprule
			$m$     & $\leq$ 0.5    & 0.6   & 0.7   & 0.8   & 0.9   & 0.99 & 0.999 & 1 \\
			\midrule
			acc (\%) & \emph{collapse} & 21.2 & 32.8 & 39.3 & 46.5 & 56.2 & 53.1 & 31.2 \\
			\bottomrule
		\end{tabular}
		\vspace{0.1cm}
		\caption{The parameter experiments of $m$ on MoCo ($\tau=0.07$).}
		\label{tab:tau007_m}
		\vspace{-0.4cm}
	\end{table}

	In the following sections, we provide thorough experiments and visualization analysis to show how the parameter $m$ affects the contrastive learning process. We attempt various $m$ for MoCo on ImageNet-100 (denoted as IN-100) \cite{tian2019contrastive} with linear readout protocol for evaluation (details in Supp \textbf{B}).
	As the Tab \ref{tab:tau007_m} shown, with the decrease of $m$ (increasing the update speed of encoder $f_k$), the accuracy presents an inverse U-shape and the max $56.24\%$ locates at $m=0.99$ and the model collapse\footnote{Model collapse means that the transfer accuracy with linear readout protocol can not achieve the accuracy of training from random initialization, \ie, $15.90\%$, indicating the negative effect brought by pre-train.} when $m\leq 0.5$. The trend of these results is similar with BYOL \cite{grill2020bootstrap}. 
	
\begin{figure}
		\centering
		\subfigure[Var of neg scores]{
			\label{fig:t007_3d_negvar}
			\includegraphics[width=0.315\linewidth,height=0.315\linewidth]{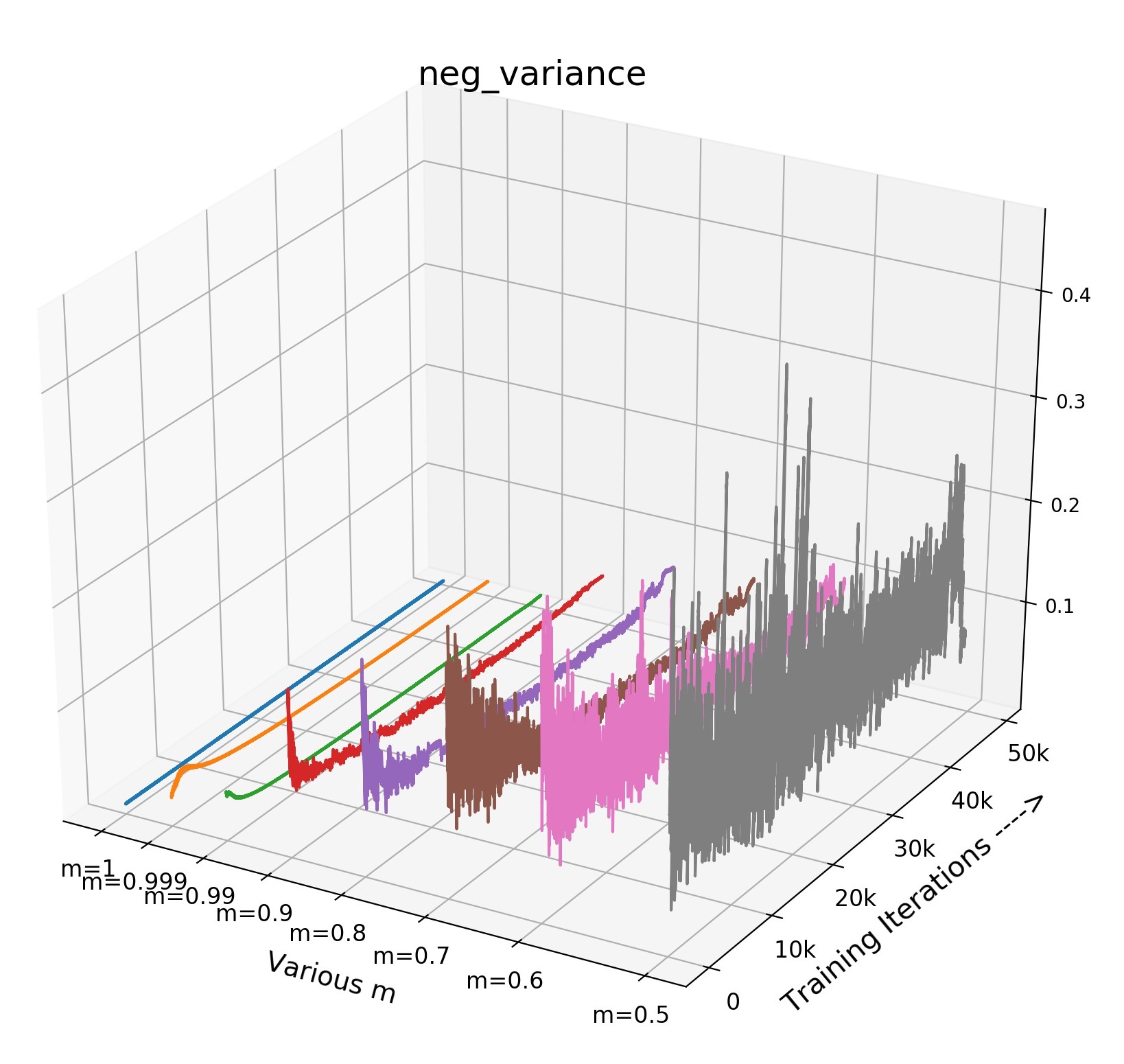}}
		\subfigure[Mean of neg scores]{
			\label{fig:t007_3d_negmean}
			\includegraphics[width=0.315\linewidth,height=0.315\linewidth]{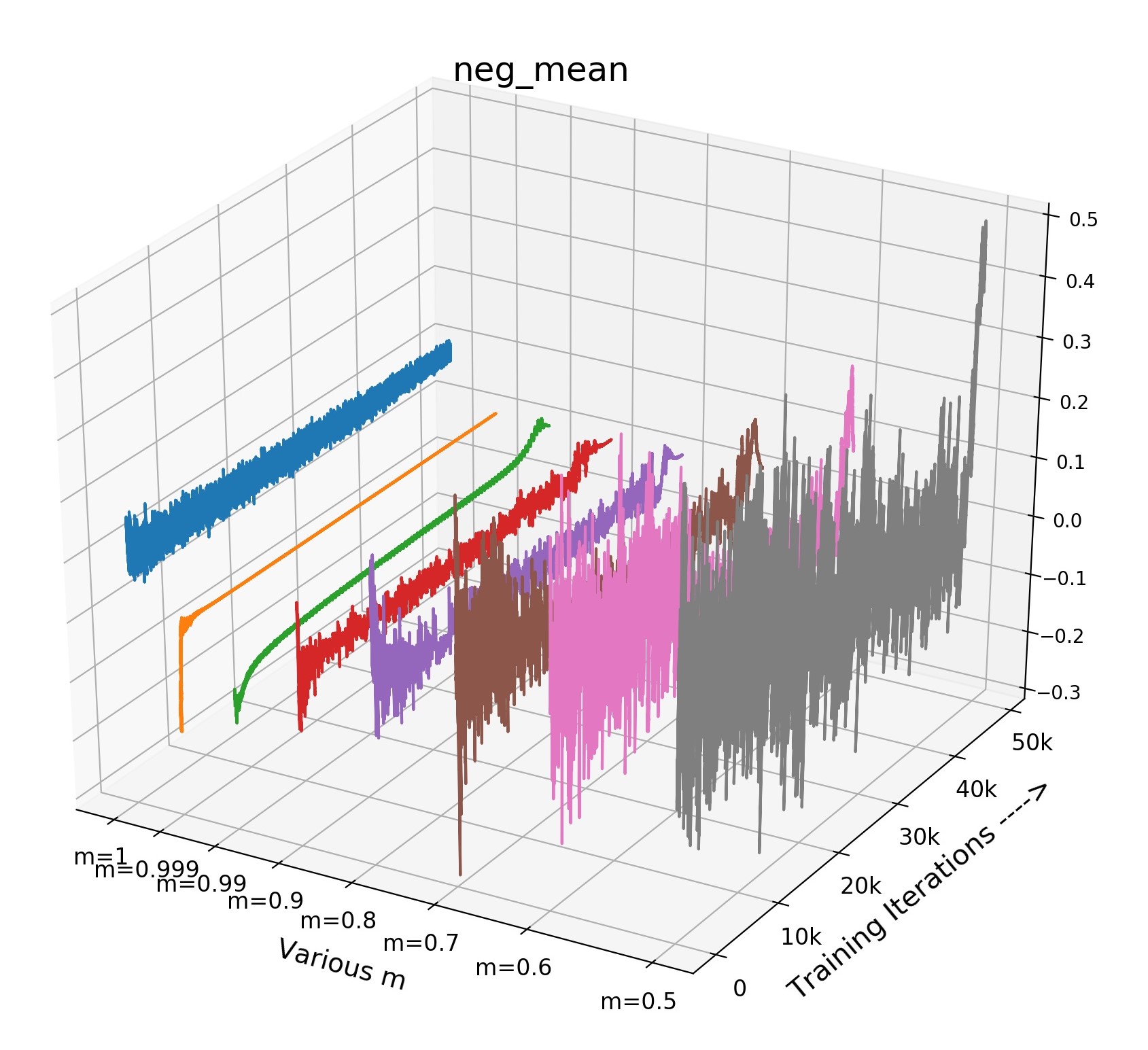}}
		\subfigure[Mean of pos scores]{
			\label{fig:t007_3d_posmean}
			\includegraphics[width=0.315\linewidth,height=0.315\linewidth]{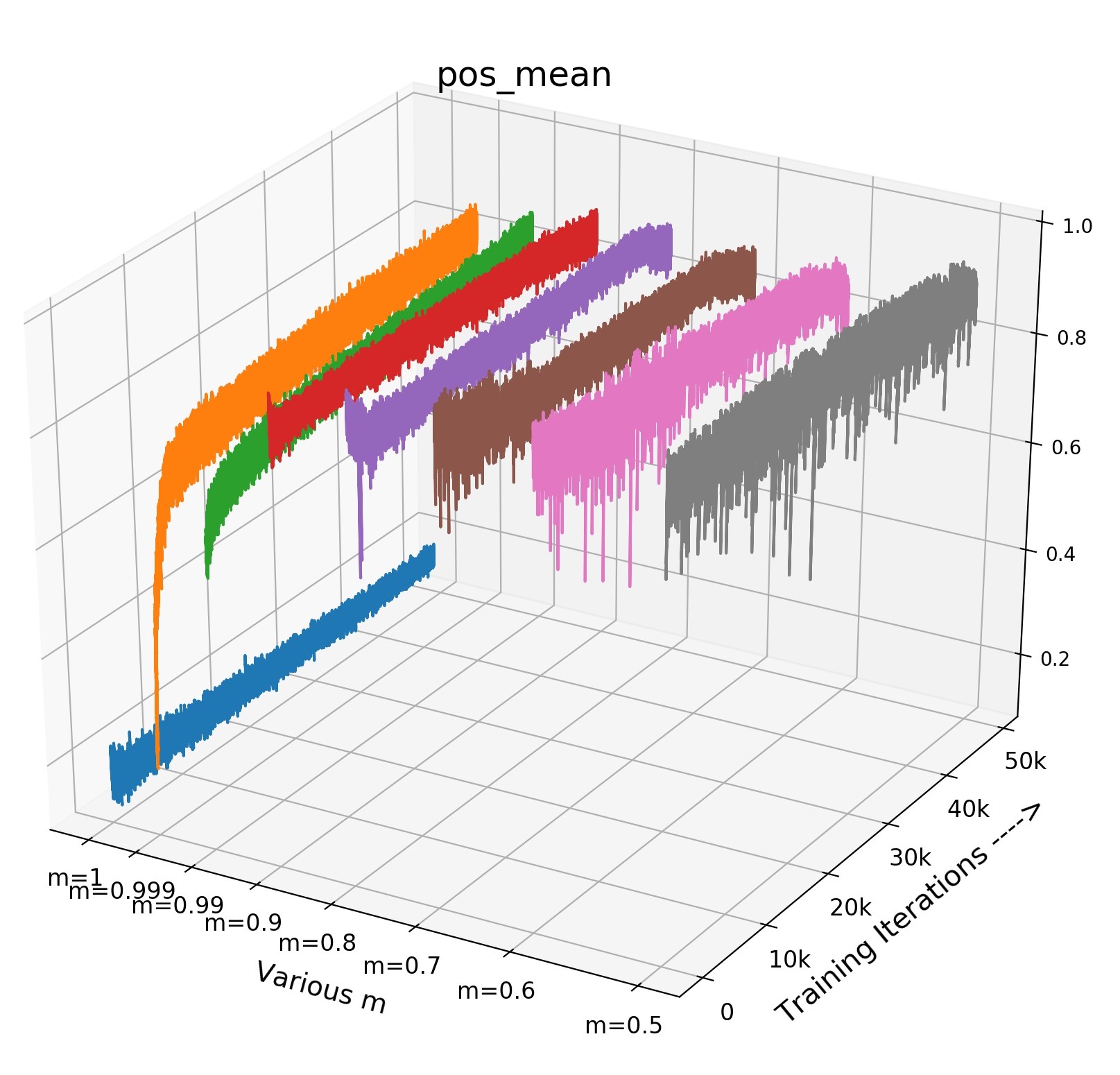}}
		%\vspace{-0.1cm}
		\caption{Pos/neg score statistics of various $m$ in MoCo training}
		\label{fig:t007}\
		\vspace{-0.5cm}
	\end{figure}

	We choose three non-trivial statistics to visualize the score distribution:
	the mean of pos/neg scores (indicating the approximate average of the pos/neg pair distance) and the variance of negative scores (indicating the fluctuation degree of the negative samples in the memory queue).
	As shown in Fig \ref{fig:t007_3d_negvar}, when $m$ becomes smaller, the update speed of encoder $k$ is increasing, leading to incremental differences of features among training steps, which is reflected as the growing variance of negative scores of the queue, namely the inconsistency. 
	Specifically, when $m=1$ (no update of $f_k$ during training), the variance is closed to zero (blue line) while the variance of $m=0.9$ (red) is larger but relatively unstable. $m=0.5$ (grey) brings more violent fluctuations/inconsistency in the memory queue, leading to a poor transfer accuracy even model collapse.

	\begin{figure}
		\centering
		\subfigure[$m=0.99$ $|$ $56.2\%$]{
			\label{fig:t007_m099_L2grad}
			\includegraphics[width=0.31\linewidth,height=0.28\linewidth]{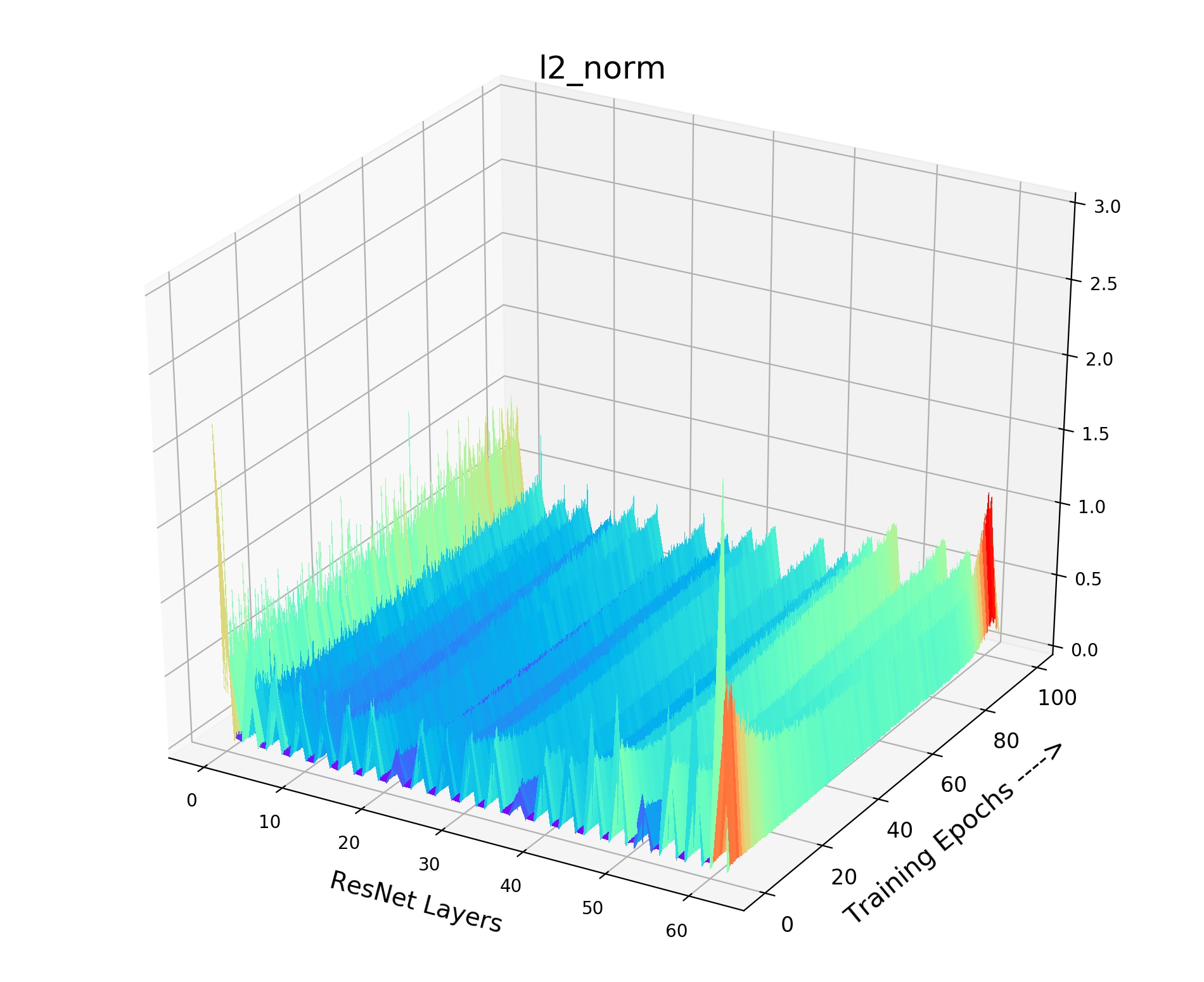}}
		\subfigure[$m=0.6$ $|$ $21.2\%$]{
			\label{fig:t007_m06_L2grad}
			\includegraphics[width=0.31\linewidth,height=0.28\linewidth]{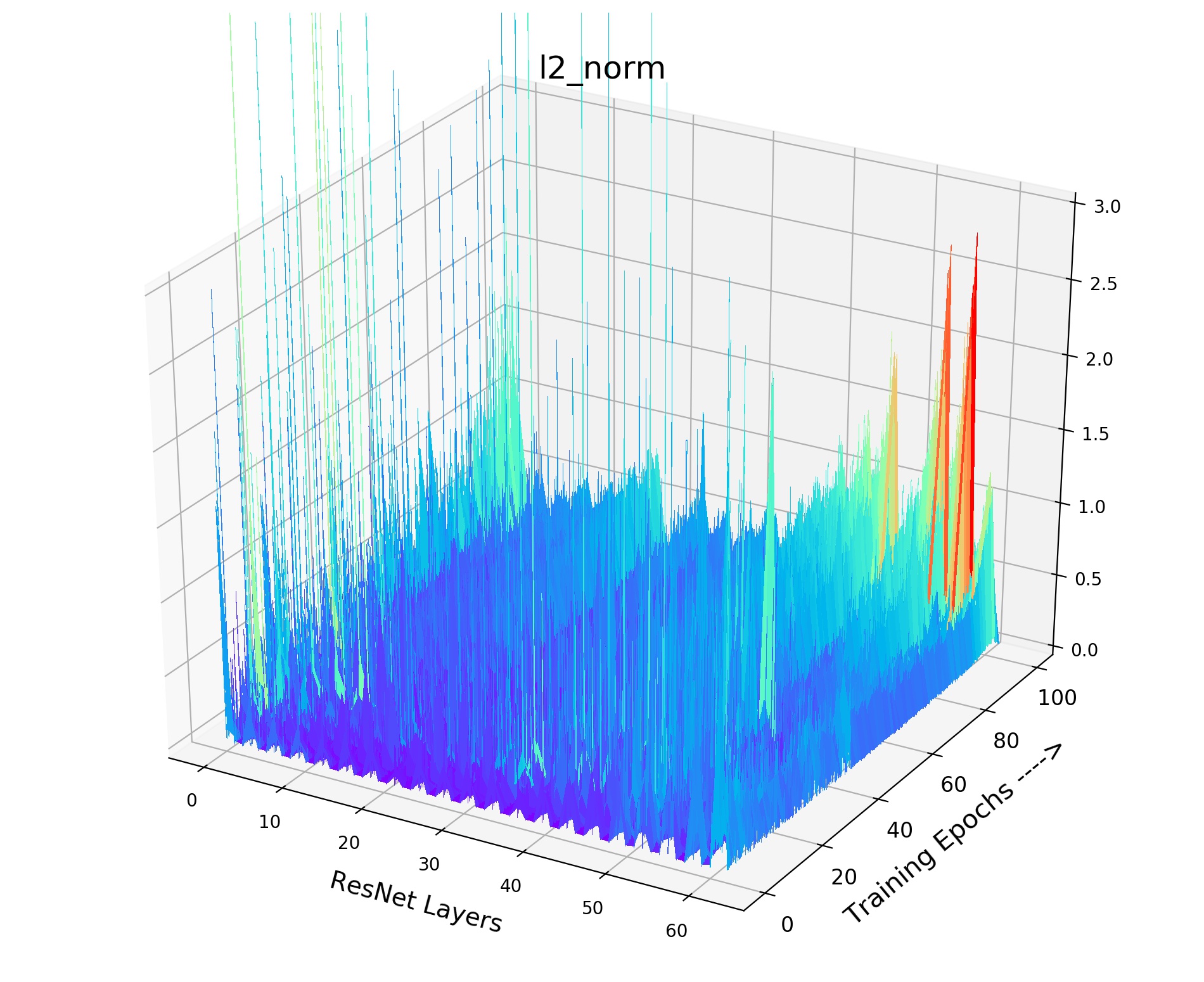}}
		\subfigure[$m=0.5$ $|$ collapse]{
			\label{fig:t007_m05_L2grad}
			\includegraphics[width=0.31\linewidth,height=0.28\linewidth]{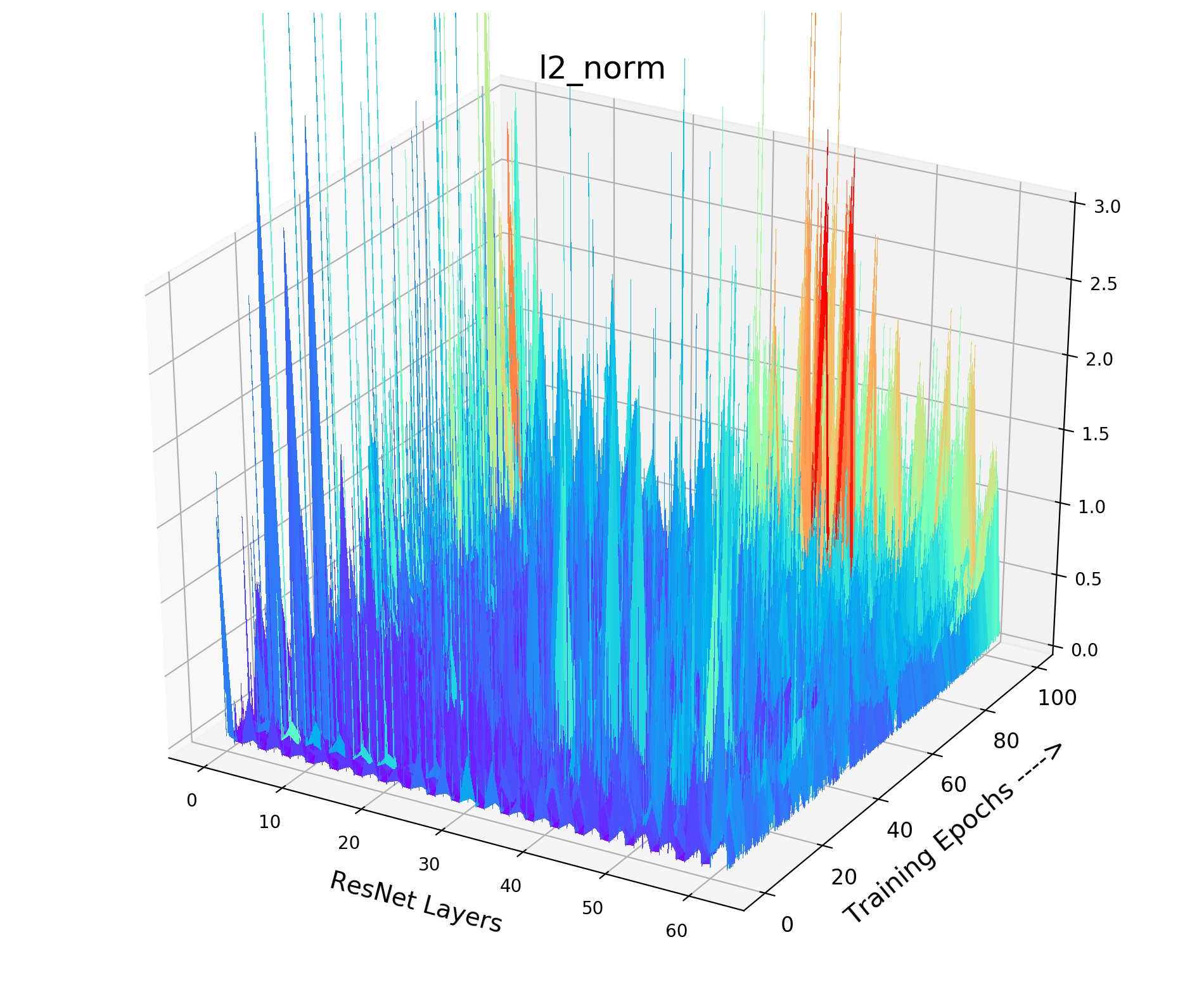}}
		%\vspace{-0.1cm}
		\caption{Gradient ($\ell_2$ norm) landscape of various $m$}
		\label{fig:t007_grad}
		\vspace{-0.3cm}
	\end{figure}

	{\noindent \textbf{Inside Analysis of Model Collapse:}}
	The model collapse is caused by various reasons. Small $m$ (fast update speed of $f_k$) brings not only the inconsistency, but also the confusion of negative scores. For the mean of neg scores (lines in Fig \ref{fig:t007_3d_negmean}), the volatility degree of $m=0.6$ (pink) and $m=0.5$ (grey) is much sharper than the best model $m=0.99$ (green).
	The mean of neg scores reflects the approximate score for all the negative pairs in the memory queue. 
	If it becomes drastically volatile with the training process, the corresponding loss value and gradient will fluctuate violently, resulting in bad convergence. As shown in Fig \ref{fig:t007_grad}, the smooth and stable gradient landscape of $m=0.99$ (Fig  \ref{fig:t007_m099_L2grad}) becomes sharp and messy with the decrease of $m$ (Fig \ref{fig:t007_m06_L2grad} for $m=0.6$ and Fig \ref{fig:t007_m05_L2grad} for $m=0.5$). Details of gradient landscape are put in Supp \textbf{C}. 
	\emph{Basically,  to learn a better pre-trained model, we need to prepare negative pairs that can maintain the stability and smoothness of score distribution and gradient for the training process, which is similar to supervised learning \cite{santurkar2018does}}.

	\begin{figure}
		\centering
		\subfigure[Var of neg scores]{
			\label{fig:t007_2d_negvar}
			\includegraphics[width=0.3\linewidth,height=0.325\linewidth]{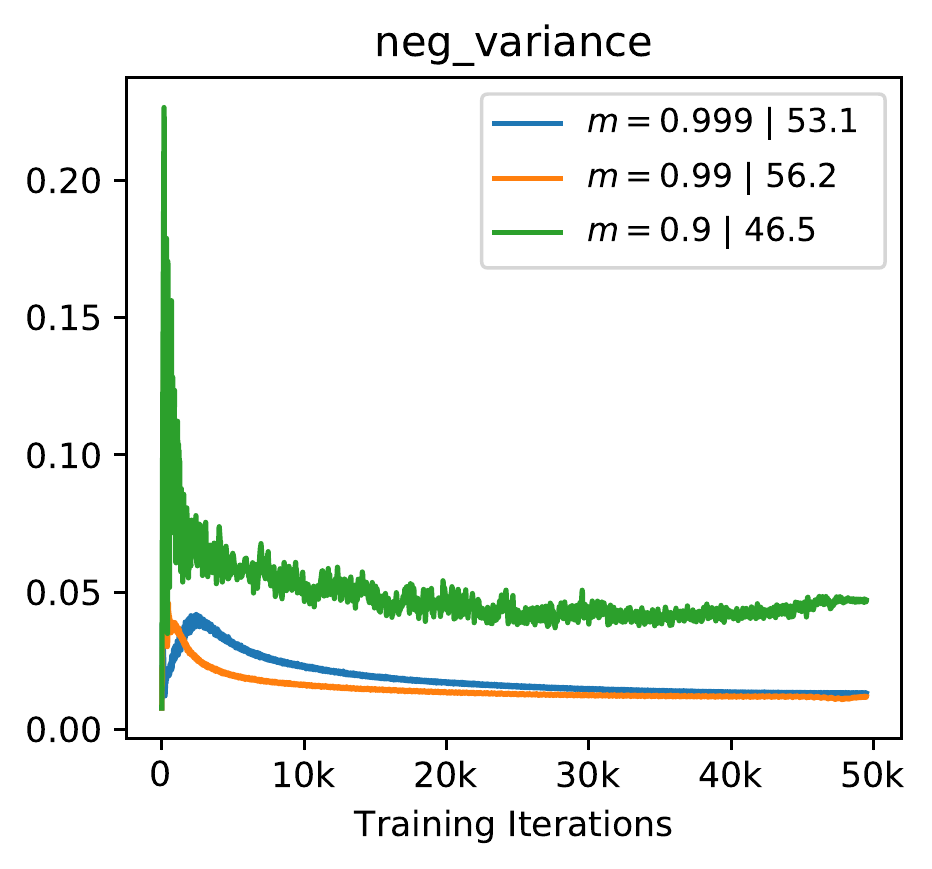}}
		\subfigure[Mean of neg scores]{
			\label{fig:t007_2d_negmean}
			\includegraphics[width=0.3\linewidth,height=0.325\linewidth]{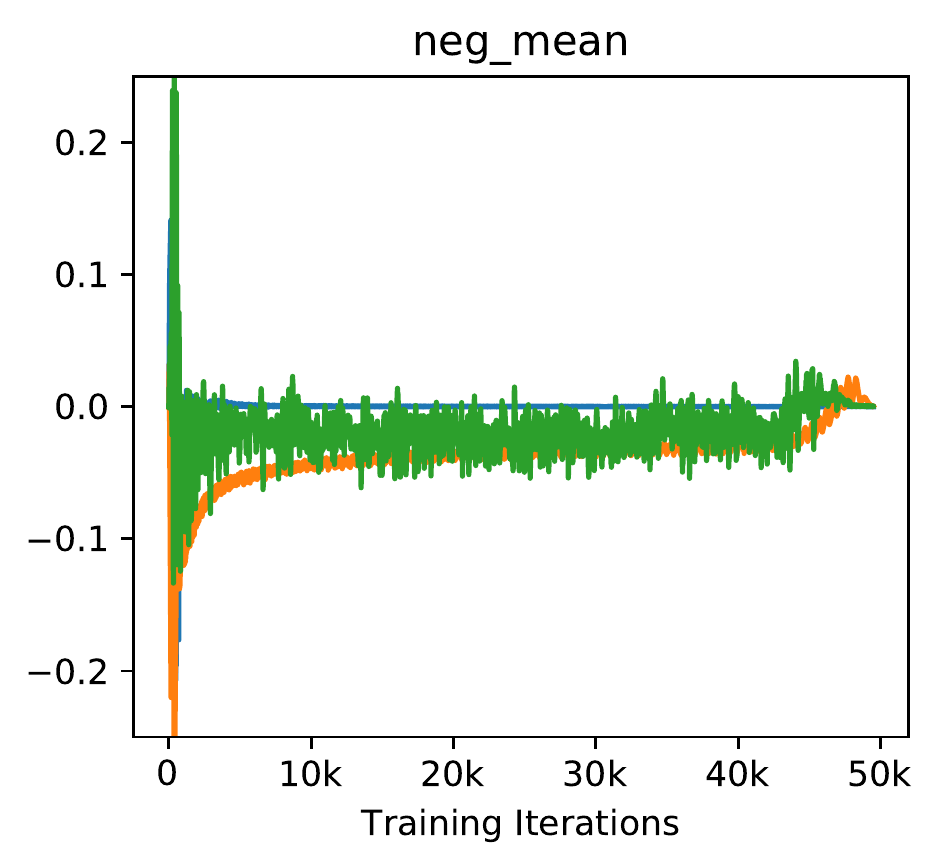}}
		\subfigure[Mean of pos scores]{
			\label{fig:t007_2d_posmean}
			\includegraphics[width=0.3\linewidth,height=0.325\linewidth]{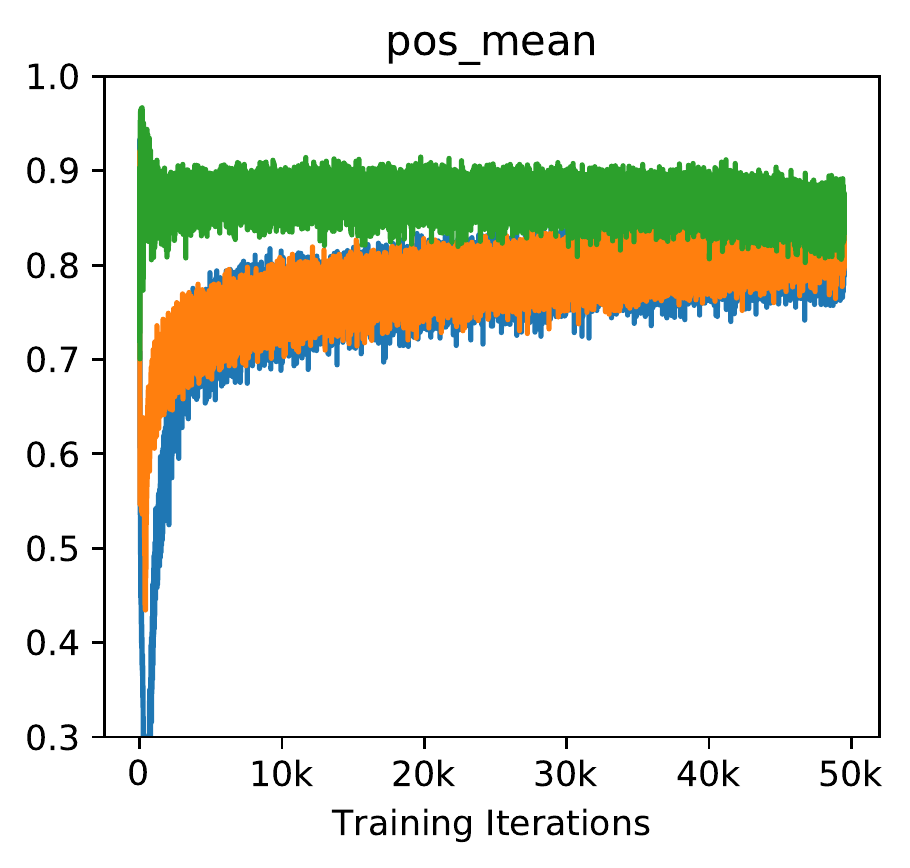}}
		\caption{2D view of pos/neg score statistics of various $m$}
		\label{fig:t007_2d}
		\vspace{-0.5cm}
	\end{figure}
	
	{\noindent \textbf{Hard Positive Boosts Performance:}}
	Small $m$ not only indicates the faster update speed, but also more similarity between encoder $f_k$ and $f_q$, \ie, in an extreme case, when $m=0$, the parameters $\theta_k$ is completely the same with $\theta_q$ in each training step.
	The increasing similarity of encoder $f_q$ and $f_k$ will reduce the dissimilarity between $z_q$ and $z_{k^{+}}$, and only the view variance brought by data augmentations remains, leading to a higher positive score. Fig \ref{fig:t007_3d_posmean} shows that high positive scores of $m\leq0.9$ will produce easy positive pairs with the close distance and little view variance in feature space. 
	
	However, in Fig \ref{fig:t007_2d_posmean}, when we increase $m$ from $0.9$ (green) to $0.99$ (orange), the easy pos pair becomes  hard pos pair (from very similar $~0.9$ to less similar $~0.7$), leading to a higher transfer accuracy ($46.5\%$ \emph{v.s} $56.2\%$, $9.7\%$ increased). 
	Note that this observation (converting easy positive to hard one) could be explained by InfoMin principle \cite{tian2020makes}: Raising the view variance between $z_q$ and $z_{k^{+}}$ corresponds to increasing the mutual information for contrastive learning, which forces the encoder learns a more robust embedding and thus improves the transfer accuracy.
	
	\emph{In the guarantee of stable and smooth score distribution and gradient, we can adopt some feature transformation methods which create hard ones by decreasing easy positive scores.}
	Thus, we propose a positive feature extrapolation method to improve transfer accuracy in section \ref{pos_exp}.

	\section{Proposed Feature Transformation Method}
\begin{figure}
		\centering
		\vspace{-0.4cm}
		\subfigure[Negative Interpolation]{
			\label{fig:inter}
			\includegraphics[width=0.45\linewidth,height=0.25\linewidth]{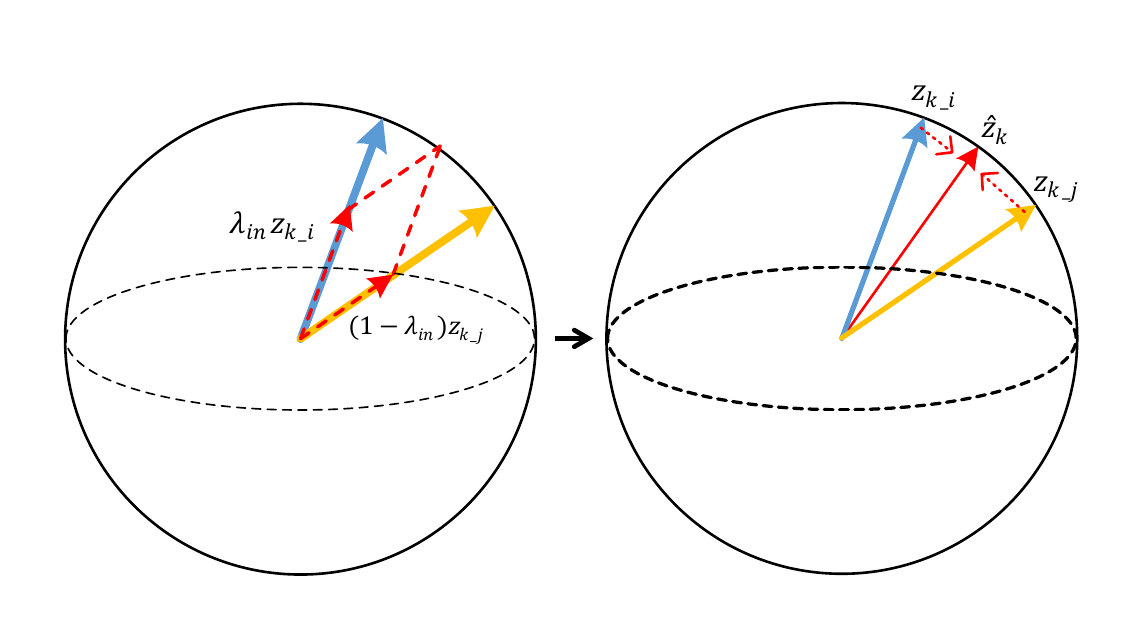}}
		\subfigure[Positive Extrapolation]{
			\label{fig:extra}
			\includegraphics[width=0.5\linewidth,height=0.29\linewidth]{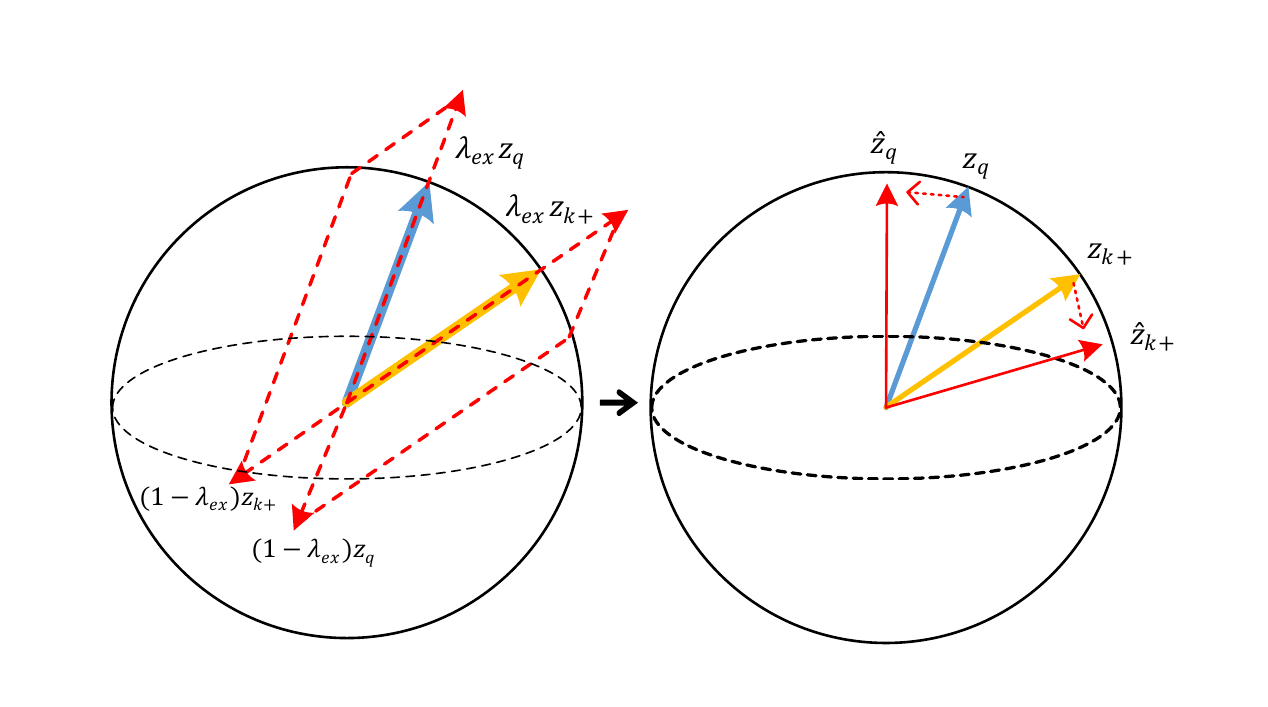}}
		%\vspace{0.1cm}
		\caption{The process of our proposed negative interpolation and positive extrapolation. For the negative interpolation, we randomly interpolate two features in memory queue to produce a new negative. For positive extrapolation, the two positive features are pushed away from each other using extrapolation, changing easy positives to hard positives, which is better for contrastive learning.}
		\label{fig:inter_and_extra}
		\vspace{-0.4cm}
	\end{figure}

	The learning objective of Info-NCE is to draw the positive pair ($z_q$ and $z_{k^{+}}$) closer while pushing away negative pairs ($z_q$ and all the $z_{k^{-}}$ in memory queue) in the embedding space. 
	Therefore, we could directly apply feature transformation on the pos/neg features, in order to provide appropriate regularization \cite{verma2019manifoldmix} or make the learning harder \cite{tian2020makes}. Specifically, we develop positive extrapolation to transform the original positive pair to be further to increase the hardness and negative interpolation of memory queue to increase the diversity of negative samples, as Fig \ref{fig:inter_and_extra} shown. Notably, our method
	does not change the loss terms because it only replaces original pair scores with the new transformed pos/neg scores for calculating loss term.

	\begin{table}
		\vspace{-.5em}
		\centering
		\small
		\begin{tabular}{c|ccccccc}
			\toprule
			$\alpha_{ex}$ & -  &  0.2    & 0.4   & 0.6   & 1.4   & 1.6   & 2.0 \\
			\midrule
			acc (\%) & 71.1 & 71.6  &71.8 & 71.9 & 72.7 & 72.4   & \textbf{72.8} \\
			\bottomrule
		\end{tabular}
		\vspace{0.1cm}
		\caption{\label{tab:in100_lambda_ex} Various $\alpha_{ex}$ for positive extrapolation, the best result is marked in bold.  We employ ResNet-50 \cite{res} for the results. '-' indicates MoCo baseline without using extrapolation.}
		\vspace{-0.4cm}
	\end{table}
	
	\subsection{Positive Extrapolation}\label{pos_exp}
	Following the discussion in Sec \ref{sec:visual_example} which indicates that lowering the easy positive pair scores to create hard positive pairs during training could be beneficial for the final transfer performance.
	Thus we would like to explore a way to manipulate the positive features $z_q$ and $z_{k^{+}}$ to increase the view variance between them during training.
	
	First, we simply adopt weighted addition for the two positive features to generate new feature:
	\begin{align}
	\hat{z}_q &= \lambda_{ex} z_q + (1 - \lambda_{ex}) z_{k^{+}} \nonumber\\
	\hat{z}_{k^{+}} &= \lambda_{ex} z_{k^{+}} + (1 - \lambda_{ex}) z_q 
	\label{equal:extrapolation} 
	\end{align}
	where $\hat{z}_q$ and $\hat{z}_{k^{+}}$ are the transformed new features. Meanwhile, considering the design principle of mixup \cite{verma2019manifoldmix, zhang2017mixup}, we make sure that the summation of weights equals to $1$. More importantly, we should guarantee than the transformed pos score $\hat{S}_{q\cdot k^{+}}$ is smaller than the original pos score $S_{q\cdot k^{+}}$, namely $\hat{z}_q\hat{z}_{k^{+}} \leq z_qz_{k^{+}}$. Take Equation \ref{equal:extrapolation} into the transformed score:
	\begin{align}
	\hat{S}_{q\cdot k^{+}} &= 2\lambda_{ex} (1 - \lambda_{ex})(1-S_{q\cdot k^{+}}) + S_{q\cdot k^{+}} \leq S_{q\cdot k^{+}}
	\label{equal:range} 
	\end{align}
	Because $S_{q\cdot k^{+}} \in [-1, 1]$ and thus $(1-S_{q\cdot k^{+}})\geq 0$. To make sure the lower score $\hat{S}_{q\cdot k^{+}} \leq S_{q\cdot k^{+}}$, we need to set  $\lambda_{ex}\geq1$ to let $2\cdot\lambda_{ex} (1 - \lambda_{ex})\leq 0$. So we choose $\lambda_{ex} \sim Beta(\alpha_{ex}, \alpha_{ex}) + 1$ \footnote{We choose to set the two parameter $\alpha_{ex}$ of the beta distribution to be the same, because the two mixed  features  are symmetrical. And the same applies to the negative feature interpolation.
	} is sampled from a beta distribution and then adding $1$ results in a range of $(1,2)$. And the range of transformed pos score will be $\hat{S}_{q\cdot k^{+}} \in [ -4+ 5S_{q\cdot k^{+}} , S_{q\cdot k^{+}}]$.
	
	Intuitively, it can be seemed as a simple approach to push away $z_q$ and $z_{k^{+}}$ in feature space. After extrapolation, the distance between the extrapolated feature vector is enlarged. Therefore the extrapolation can serve as a feature transformation to create hard positives from easy ones. As shown in Fig \ref{fig:extra}, it brings a minor direction change for two positive vectors and meanwhile conveying a larger view variance of a sample for better contrastive learning. The visualization of lowering pos score by extrapolation is shown in Fig~\ref{fig:performance_gain}.	
	
	We evaluate the efficacy of positive extrapolation on IN-100 and attempt various $\alpha_{ex}$ in Tab \ref{tab:in100_lambda_ex}.
	The positive extrapolation with various $\alpha_{ex}$ consistently improves the accuracy from the baseline MoCo ($71.1\%$), which clearly demonstrates the efficacy of positive extrapolation. It is interesting that $\alpha_{ex}\textgreater 1$ will get better results than those of $\alpha_{ex} \textless1$. Because the beta distribution with $\alpha_{ex}\textless 1$ provides extreme large or small $\lambda_{ex}$ with high probability, \eg, $1.1$ or $1.9$, while the beta distribution with $\alpha_{ex}\textgreater1$ gives neutral  $\lambda_{ex}=1.5$ with high probability \footnote{The beta distribution with $\alpha_{ex}\textgreater1$ shows an inverted U shape which samples 0.5 with a greater probability and thus making $\lambda_{ex}$ to have a greater chance to be $1.5$.}.
	According to Equation \ref{equal:range}, extreme $\lambda_{ex}$ will bring too much/little hardness, so the corresponding performance is not robust as the neutral one.

	\begin{table}[]
		
		\begin{tabular}{@{}lcc@{}}
			\toprule
			Method & $\alpha_{ex}$ & pos interpolation/extrapolation    \\
			\midrule
			MoCo   & 0.2 &    69.1 / 71.6 \\
			(baseline: 71.1 )       &  2.0 &  67.4 / 72.8        \\
			\bottomrule
		\end{tabular}
		\vspace{0.1cm}
		\caption{\label{tab:in100_lambda_comp} Positive extrapolation v.s. interpolation. Interpolation hurts the performance while extrapolation improves.}
		\vspace{-0.4cm}
	\end{table}

	{\noindent \textbf{What if Positive Interpolation?}}
	To further verify our conjecture that extrapolation can create hard positives while interpolation won't,  we also conduct experiments for the interpolation of positive features, shown in Tab~\ref{tab:in100_lambda_comp}. We can observe a clear performance drop ($5.4\%$ drop for  neutral $\alpha_{ex}=2$) for this experiment. The reason is that the interpolation between positive features pulls the positive pairs together thus reducing the hardness in the training process. In other words, the view variance of positive pairs is decreasing, and thus easy to cause non-robust features.

	\subsection{Negative Interpolation}
	Previous contrastive models \cite{chen2020simple,he2020momentum} do not make full use of negative samples. 
	\eg, In MoCo, there are many repetitive negative features stores in the memory queue iteration by iteration. 
	Thus we could design a new strategy to fully utilize negative features and increase the diversity of the memory queue. With sufficient randomness, 
	we propose the negative interpolation in memory queue, which intuitively provides diversified negatives  for each training step.
	
	Specifically, we denote the negative memory queue of MoCo as $Z_{neg}=\{z_{1},z_{2}, \dots, z_{K}\}$ where $K$ is the size of the memory queue, and $Z_{perm}$ as the random permutation of $Z_{neg}$.
	We propose to use a simple interpolation between two memory queue to create a new queue $\hat{Z}_{neg}=\{\hat{z}_{1}, \hat{z}_{2}, \dots, \hat{z}_{K}\}$:
	\begin{equation}
	\hat{Z}_{neg} = \lambda_{in}\cdot Z_{neg} + (1 - \lambda_{in})\cdot Z_{perm}
	\end{equation}
	where $\lambda_{in} \sim Beta(\alpha_{in}, \alpha_{in})$ is in the range of $(0,1)$, as Fig \ref{fig:inter} shown. The transformed memory queue $\hat{Z}_{neg}$ provides fresh interpolated negatives for contrastive loss iteration by iteration,
	where the random permutation and $\lambda_{in}$  ensure the diversity of $\hat{Z}_{neg}$ of each training step. 
	The diversity makes the model to compare with much more linear combinations of previous negatives in each training step. 
	Positive extrapolation increases the view variance between two pos features while the negative interpolation similarly boosts the ``sample variance'' (diversity) of the memory queue. 
	We conjecture that original queue $Z_{neg}$ provides discrete distribution of negative samples but our method can fill in the incomplete sample points of the distribution by random interpolation, leading to a more discriminative model
	We evaluate the efficacy of negative interpolation on IN-100 and attempt various $\alpha_{in}$ in Tab \ref{tab:in100_lambda_in}.
	The neg interpolation is fairly robust with various $\alpha_{in}$, with the improvement of $2.2\%$-$3.5\%$ from the baseline ($71.1\%$). More interesting discussions about negative feature transformation (hard negatives \& negative extrapolation) are shown in Supp \textbf{G}.

	\begin{table}
		\vspace{-.5em}
		\centering
		\small
		\begin{tabular}{c|ccccccc}
			\toprule
			$\alpha_{in}$ & -  &  0.2    & 0.4   & 0.6   & 1.4   & 1.6   & 2.0 \\
			\midrule
			acc (\%) & 71.1 & 73.3  &74.1 & 74.2 & 73.5 & \textbf{74.6}   & 74.1 \\
			\bottomrule
		\end{tabular}
		\vspace{0.1cm}
		\caption{\label{tab:in100_lambda_in} Various $\alpha_{in}$ for negative interpolation, the best result is marked in bold.  We employ ResNet-50 \cite{res} for the results. '-' indicates MoCo baseline without using negative interpolation.}
		\vspace{-0.2cm}
	\end{table}

	Previous works have explored the method leveraging image-level~\cite{shen2020mix} and feature-level~\cite{kalantidis2020hard_mochi} mixing in contrastive learning.
	Our method differs from the previous works in three ways, first is the motivation, we are motivated by our observation in Sec~\ref{sec:score_vis} to propose the feature transformation strategies. 
	Second, the way we extrapolate between two positive features is novel and outperforms the other two methods on several experiments in Tab \ref{tab:cls} and \ref{tab:final}.
	Third, the negative interpolation aims at fully utilizing negative samples in  each training step. Both FT methods focus on exploring an effective  way to perform feature transformation, not simply extending hard negatives to memory queue \cite{kalantidis2020hard_mochi}, neither the image-level mixup \cite{shen2020mix}. 
	In the following sections,
	we provide inside discussions for the proposed FT, including (1) What if extending memory queue instead of FT. (2) When to add FT? (3) Dimension-level mixing rather than linear mixup. (4) Could the gains brought by FT vanish if training longer?

	\subsection{Discussions}

	\begin{table}[]
		\centering
		\setlength\tabcolsep{3pt}
		\begin{tabular}{@{}lcccc@{}}
			\toprule
			Method & $\alpha_{in}$   & $Z_n$& queue size & Acc   \\
			\midrule
			moco+  original queue                        & -    & $Z_{neg}$ & $K$ & 71.10 \\
			moco+  original queue                        & -    & $Z_{neg}$ & $2K$ & 71.40 \\
			moco+ Neg FT queue      &  1.6 & $\hat{Z}_{neg}$  &$K$  &  74.64   \\
			moco+ Neg FT+original       &  1.6 & $\tilde{Z}_{neg}$   & $2K$  &  74.73   \\
			\bottomrule
		\end{tabular}
		\vspace{0.1cm}
		\caption{\label{tab:in100_lambda_in_concat} Ablation results for using different queue of negative features (Res50). The transformed queue $\hat{Z}_{neg}$  can completely replace the extended queue $\tilde{Z}_{neg}$ with small computations.}
		\vspace{-0.4cm}
	\end{table}
	
	\begin{figure}
		\vspace{-0.4cm}
		\centering
		\subfigure[Mean of neg scores]{
			\label{fig:whentoadd_negmean}
			\includegraphics[width=0.4\linewidth,height=0.35\linewidth]{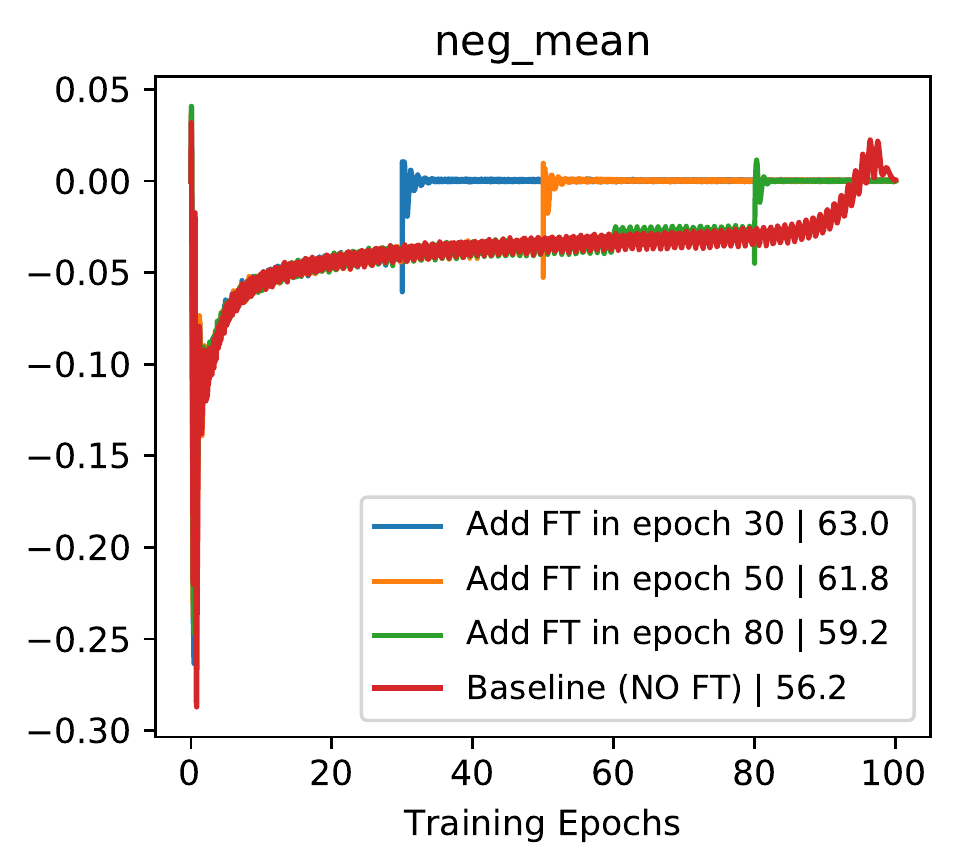}}
		\subfigure[Mean of pos scores]{
			\label{fig:whentoadd_posmean}
			\includegraphics[width=0.4\linewidth,height=0.35\linewidth]{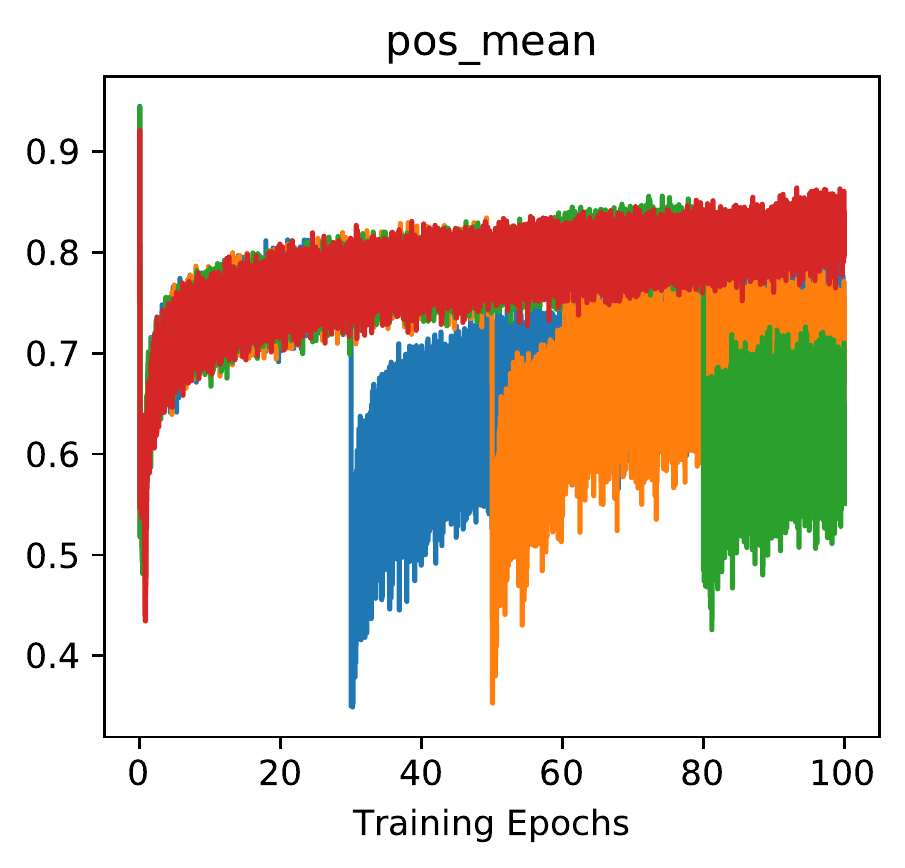}}
		\subfigure[Baseline MoCo landscape]{
			\label{fig:whentoadd_baseline}
			\includegraphics[width=0.45\linewidth,height=0.4\linewidth]{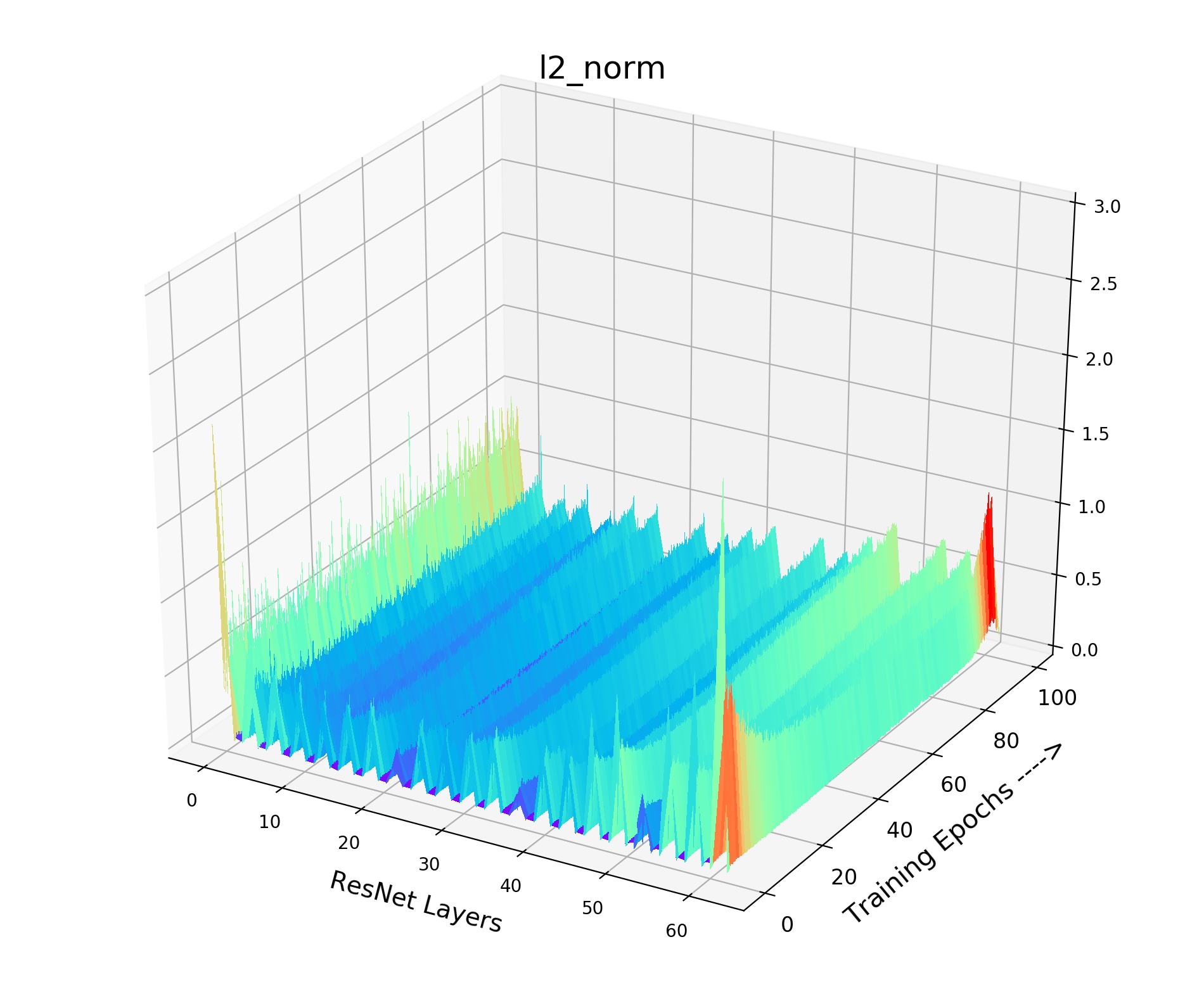}}
		\subfigure[Adding FT in 50th epoch]{
			\label{fig:whentoadd_50ep}
			\includegraphics[width=0.45\linewidth,height=0.4\linewidth]{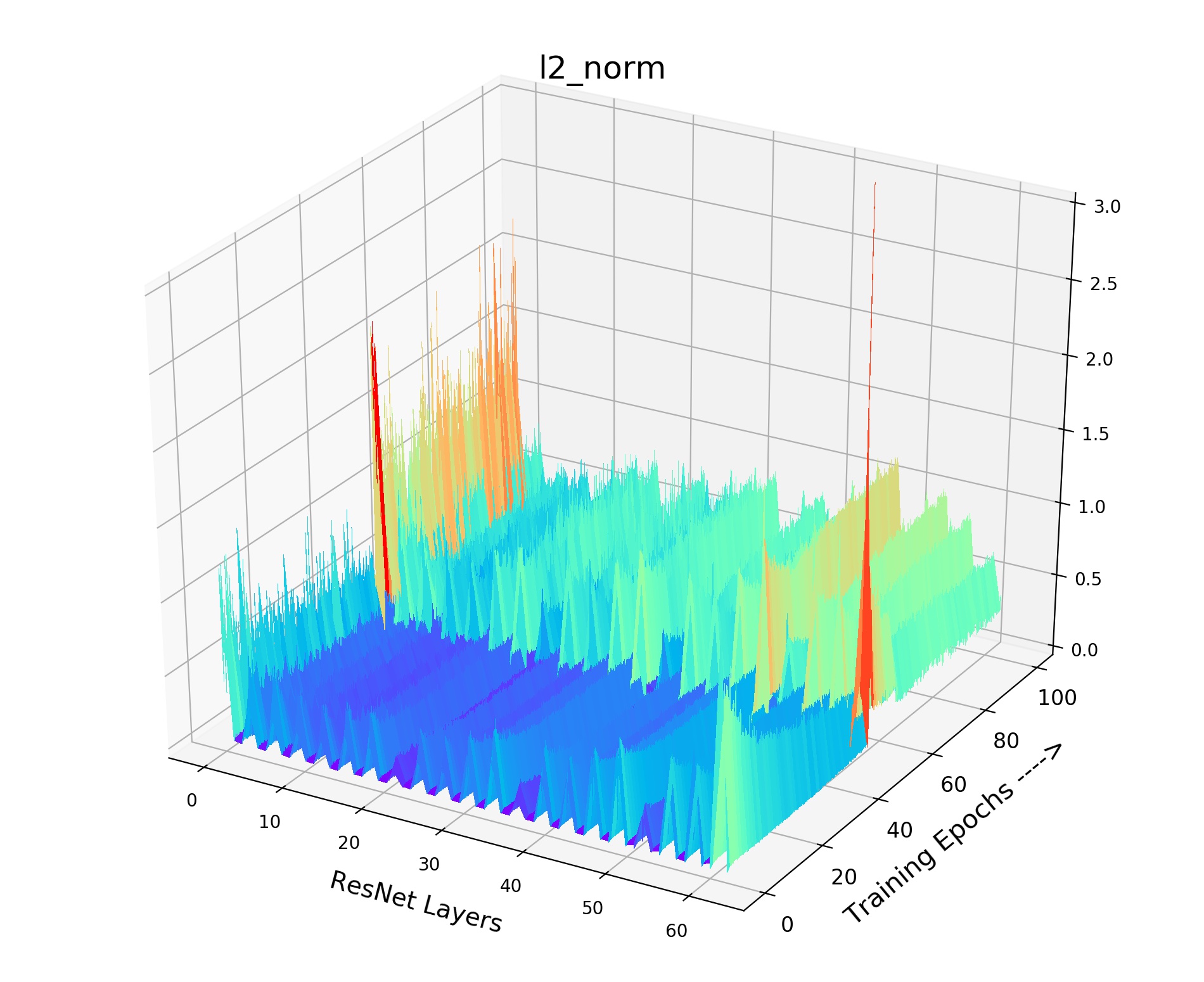}}
		%\vspace{-0.1cm}
		\caption{Visualization of when to add FT, including score distribution and Gradient ($\ell_2$ norm) landscape.}
		\label{fig:whentoadd}
		\vspace{-0.1cm}
	\end{figure}
	
	{\noindent \textbf{Extending memory queue instead of FT: }}
	Previous works~\cite{he2020momentum,chen2020simple} show that increasing the number of negative example ($K$) in contrastive learning could be beneficial for the final performance, thus they either uses a memory queue~\cite{he2020momentum} or a large  batchsize~\cite{chen2020simple} to obtain more negative examples.
	Specifically, \cite{oord2018representation, hjelm2018learning, tian2020makes} shows that increasing $K$ will improve the lower bound of the mutual information. The negative interpolation method could also be leveraged to enlarge the number of negative examples: 
	We use the union queue of original negatives and interpolated queue, $\tilde{Z}_{neg}=\hat{Z}_{neg}\cup Z_{neg}$, which contains twice the number of negative examples ($2K$) than $\hat{Z}_{neg}$. 
	
	We compare the performance of using only the interpolated queue $\hat{Z}_{neg}$, original  $Z_{neg}$ with $K$/$2K$ negative samples, and their combination $\tilde{Z}_{neg}$, in Tab~\ref{tab:in100_lambda_in_concat}.
	We found that using the combination queue shows negligible improvement over the performance ($74.73\%$) of using the interpolated queue alone ($74.64\%$). 
	We consider that the interpolated negative features contain  sufficient diversified negatives compared with the original queue. So even the double negative samples (more mutual information) of the extended queue ($\tilde{Z}_{neg}$) cannot boost the performance. %Mochi\cite{kalantidis2020hard_mochi} presents the similar trend that adding hard negatives to the original bank will bring very slight improvement and even hurt the performance in their original paper and Tab \ref{tab:in1k_moco}.
	Notably, the extended queue requires double times computation for each contrastive loss. Thus we recommend feature transformations with less computation but more efficacy rather than feature augmentation.

	\begin{table}
		\vspace{-.5em}
		\centering
		\small
		\begin{tabular}{c|cccccc}
			\toprule
			FT begin epoch & 0  &  2    & 30   & 50   & 80   & -   \\
			\midrule
			Res18 acc (\%) & 62.6 & \textbf{63.3}  & 62.9 & 61.8 & 59.2 & 56.2    \\
			Res50 acc (\%) & \textbf{76.9} & 76.4  & 75.9 & 74.0 & 72.2 & 71.1    \\
			\bottomrule
		\end{tabular}
		\vspace{0.05cm}
		\caption{\label{tab:in100_whentoadd} When to add feature transformation.  We employ Res-18 (total 100 epochs) and Res-50 (total 200 epochs) on IN-100 for the results. '-' indicates MoCo baseline without using any FT.}
		\vspace{-0.5cm}
	\end{table}
	
	{\noindent \textbf{When to add feature transformation? }}
	Here we present the efficacy of FT by analysis of starting FT in various training stages. As shown in Tab \ref{tab:in100_whentoadd}, starting FT (pos extrapolation + neg interpolation) from various epoch can consistently boost the accuracy of baseline, and starting from earlier can improve more ($7.1\%$/$5.8\%$ boosts with Res-18/Res-50). With the visualizations of score distribution and gradient landscape in Fig \ref{fig:whentoadd}, we can see that our FT brings hard positives (lowering pos scores in Fig \ref{fig:whentoadd_posmean}) and hard negatives (rising neg scores in Fig \ref{fig:whentoadd_negmean}) simultaneously when the combined FT is inserted in various stages. Besides, with the comparison of the gradient ($\ell_2$ norm) landscape, we can observe that our FT brings a greater gradient for the training, which makes the model  escape from the local minima and avoid over-fitting. These analyses indicate our FT is a plug-and-play method and brings persistent view-invariance and discrimination for the training of contrastive models. More detailed discussions and visualizations are put in Supp \textbf{D}.

	{\noindent \textbf{How about Dimension-level mixing:}}
	Besides the proposed linear feature interpolation and extrapolation on the feature-level (128-d vector), we also extend the transformation to a dimension-level where the parameter $\lambda$ is a vector rather than a scalar number, this dimension-level mixing can be described as follows:
	\begin{align}
	\hat{z}_{new} &= \lambda \odot z_{i} + (1 - \lambda) \odot{ z_{j}}
	\end{align}
	where $\odot$ stands for Hadamard product, and $\lambda \in [0,1]^n$ is a vector with the same dimension as the feature vector.  
	The value of each dimension of $\lambda$ is randomly sampled from a beta distribution $\lambda_{i} \sim Beta(\alpha, \alpha)$. 
	This formulation is used for negative interpolation; For positive, $\lambda$ is added  $1$ to perform extrapolation. 	
	For neg/pos features, the dimension-level mixing could introduce more diversity/more view variance (hardness) because every dimension is performed with transformation.
	Experiments of dimension-level mixing on IN-100 shows improvement over the feature-level mixing (the 5th row in Tab \ref{tab:in100_moco}).
	
	{\noindent \textbf{Could the gains brought by FT vanish if training longer?}} Simply training longer leads to significant performance boost for contrastive pre-train. So here
	we provide the results of MoCov2/MoCov2+FT (500 epoch) on IN-100: 80.7\%-$\textgreater$81.5\%. 
	Compared with 200 epoch results (75.6\%-$\textgreater$78.3\% in Tab \ref{tab:in100_moco}), longer training actually minimizes the improvement over the baseline. 
	More training epochs can lead to comparing much more pos/neg pairs to increase the diversity. 
	However, our proposed FT accelerates this process by providing diversity and results in fast convergence, which responds to the motivation of learning diversified and discriminative representations.

	\begin{table}[]
		\centering
		\setlength\tabcolsep{1pt}
		\begin{tabular}{@{}lcccccc@{}}
			\toprule
			Method &  MoCov1 & MoCov2 & simCLR & Infomin&swav&SimSiam \\
			\midrule
			baseline$^*$   & 71.10 & 75.61 &  74.32  &81.9&82.1&77.1 \\
			+pos FT       &  72.80 & 76.22  &  75.80  & -&-&77.8 \\
			+neg FT       &  74.64  & 77.12  &  76.71 &  - &-&\\
			+both                    &  76.87  & 78.33 &  78.25  & 83.2 & 83.2&\\
			+both$_{dim}$      &  \textbf{77.21} & \textbf{79.21}  &    \textbf{78.81}  & -&-&\\
			\bottomrule
		\end{tabular}
		\vspace{0.1cm}
		\caption{\label{tab:in100_moco} Ablation studies of proposed methods on various contrastive models. The models are pre-trained for 200 epochs with Res50 on IN-100. 
			$^*$ indicates reproduced baseline results.}
		\vspace{-0.3cm}
	\end{table}

	\begin{table}[]
		\centering
		\setlength\tabcolsep{1pt}
		\begin{tabular}{@{}lllll@{}}
			\toprule
			pre-train   & IN-1k & inat-18 & CUB200  & FGVC-aircraft \\
			\midrule
			supervised  & 76.1    & 66.1 & 81.9$^*$    & 82.6$^*$        \\
			\midrule
			mocov1\cite{he2020momentum}    &  60.6   & 65.6 & 82.8$^*$   & 83.5$^*$         \\
			mocov1+ours   & 61.9    & 67.3 &83.2& 84.0\\
			\midrule
			mocov2\cite{chen2020improved}   & 67.5    & 66.8$^*$ & 82.9$^*$    & 83.6$^*$         \\
			mocov2+ours   & \textbf{69.6}    & \textbf{67.7} &\textbf{83.1}&\textbf{84.1}\\
			mocov2+MoCHi\cite{kalantidis2020hard_mochi}  & 68.0    & - & -    & -         \\
			mocov2+UnMix\cite{shen2020mix}  & 68.6    & - & -    & -         \\
			
			\bottomrule
		\end{tabular}
		\vspace{0.1cm}
		\caption{\label{tab:cls} Classification results. $^*$ indicates our reproduced results.}
		\vspace{-0.4cm}
	\end{table}
	
	\section{Experiments}
	In this section, we evaluate our Feature Transformation methods from four perspectives: (1) Ablation studies (2) FT on various contrastive models. (3) Evaluating the representation on ImageNet-1k. (4) Finetuning on various downstream tasks. 
	We keep the fairness of the experiments, especially when compared with other methods. Notice that the data augmentations are followed with the baseline methods. Details of experiments and datasets are put in Supp \textbf{B}.

	\begin{table*}[]
		\centering
		\begin{tabular}{l|c|c c c |ccc| ccc }
			\toprule
			\multirow{2}{*}{pre-train}& IN-1k &\multicolumn{3}{c|}{Faster \cite{ren2015faster_rcnn} R50-C4 VOC}  & \multicolumn{6}{c}{Mask R-CNN \cite{he2017mask} R50-C4 COCO} \\ 
			& Top-1 & AP   & AP$_{50}$   & AP$_{75}$ & AP$^{bb}$ & AP$^{bb}_{50}$ & AP$^{bb}_{75}$ & AP$^{mk}$ & AP$^{mk}_{50}$ & AP$^{mk}_{75}$ \\
			\midrule
			random init$^*$&- 		& 33.8  & 60.2  & 33.1   & 26.4 & 44.0  & 27.8  & 29.3  & 46.9 & 30.8  \\
			supervised$^*$ & 76.1	& 53.5 	& 81.3  & 58.8   & 38.2 & 58.2  & 41.2  & 33.3  & 54.7 & 35.2  \\
			infomin$^*$    & 70.1  	& 57.6	& 82.7  & 64.6   & 39.0 & 58.5  & 42.0  & 34.1  & 55.2 & 36.3  \\
			\midrule
			mocoV1\cite{he2020momentum}    & 60.6 	& 55.9  & 81.5  & 62.6   & 38.5 & 58.3  & 41.6  & 33.6  & 54.8 & 35.6  \\
			mocoV1+ours   & 61.9 	& 56.1  & 82.0	& 62.0   & 39.0 & 58.7  &42.1   & 34.1  & 55.1  & 36.0  \\
			\midrule
			mocoV2\cite{chen2020improved}    & 67.5 	& 57.0  & 82.4 	& 63.6   & 39.0 & 58.6  & 41.9  & 34.2  & 55.4 & 36.2  \\
			\textbf{mocoV2+ours}   & \textbf{69.6} 	& \textbf{58.1}  & \textbf{83.3}	& 65.1   & \textbf{39.5} & \textbf{59.2}  & 42.1  & \textbf{34.6}  & 55.6 & 36.5  \\
			mocoV2+mochi\cite{kalantidis2020hard_mochi}  & 68.0   & 57.1  & 82.7  & 64.1   & 39.4 & 59.0  &42.7   & 34.5  & 55.7 & 36.7\\
			%moco-V2+UnMix  & 68.6   & 57.7  & 83.0  & 64.3   & -    &-      & -      & -     &-     &-\\
			\midrule
			DetCo\cite{xie2021detco}         & 68.6   & 57.8  & 82.6  & 64.2   & 39.4 & 59.2  &42.3   & 34.4  & 55.7 & 36.6\\
			InsLoc\cite{yang2021instance}          & -   & 57.9  & 82.9  & 65.3   & 39.5 & 59.1  &\textbf{42.7}   & 34.5  & 56.0 & 36.8\\
			\bottomrule
		\end{tabular}
		\vspace{0.1cm}
		\caption{\label{tab:final} Object detection. All model are pre-trained for 200 epochs on ImageNet-1k. $^*$  means that the results are followed from respective papers~\cite{he2020momentum,tian2020makes}. The COCO results of mocoV2 is from \cite{kalantidis2020hard_mochi}. Our results are reported using the average of 5 runs.}
		\vspace{-0.4cm}
	\end{table*}

	\subsection{Ablation study}
	We adopt the linear readout protocol \cite{he2020momentum} to compare performance for image classification on IN-100, where we freeze the features and train a supervised linear classifier using softmax. 
	Tab~\ref{tab:in100_moco} summarizes the results of ablation studies.
	We observe that the positive extrapolation and negative interpolation components are complementary which can improve the top-1 accuracy by 5.77\%/2.72$\%$ when combined on MoCoV1/MoCoV2. The dimension-level mix also shows improvement based on the already high performance of both components. The performance-boosting of ablation studies over MoCo shows the efficacy of our FT. 
	Notice that the transformed features are not necessarily on the unit sphere (\ie, has a norm of 1), we did not need to re-perform $\ell_2$ norm for transformed features, because the performance difference is negligible ($76.87\%$ \emph{v.s} post-norm $76.68\%$). More discussions about $\ell_2$ for vector length are put in Supp \textbf{F}. Here we strongly recommend to re-perform $\ell_2$ norm for the transformed features on all the datasets, for the sake of contrasting all the scores on the unit-sphere.

	\subsection{FT on various contrastive models}
	We apply our FT to various contrastive models in  Tab~\ref{tab:in100_moco}.
	It presents that our FT brings $5.77\%$, $3.93\%$ , $1.3\%$, $1.1\%$, and $0.7\%$ improvement over
	MoCo \cite{he2020momentum}, SimCLR \cite{chen2020simple}, InfoMin \cite{tian2020makes}, SWAV \cite{caron2020unsupervised} and SimSiam \cite{chen2020exploring}, respectively on IN-100 dataset (200 epoch). It is worthy to point out that the series of ablation studies of our FT can boosts the SimCLR model. The experiments shows our FT is generic and robust for various contrastive models. %Implement details in  

	\subsection{Evaluating the representation on ImageNet-1k}
	After ablations on IN-100 dataset, we use the best settings of $\alpha_{in}$ and $\alpha_{ex}$ to train a model on ImageNet-1k (IN-1K).
	Note that the dimension-level mix is not used for the experiments on IN-1K due to computational constraints.
	We apply our method on the baseline MoCo \cite{he2020momentum} and MoCoV2 \cite{chen2020improved}, which are both trained on IN-1K with 200 epochs.
	The results and comparison are summarized in Tab \ref{tab:cls}. 
	Our method improves MoCoV1 and MoCoV2 by 1.3\% and 2.1\% on Top-1 accuracy respectively which are significant on a large dataset like IN-1K. 
	UnMix~\cite{shen2020mix} and MoCHi~\cite{kalantidis2020hard_mochi} are the methods that also leverage mixup to better aid the contrastive learning process.
	Notably, we can observe that our method with MoCoV2 can provide larger performance gain than UnMix and MoCHi respectively.

	\subsection{Downstream Tasks}
	{\noindent \textbf{Fine-grained image classification}}
	We evaluate the efficacy on real world fine-grained classification datasets, \eg, large scale long-tail iNaturalist2018 \cite{van2018inaturalist},  CUB-200 \cite{welinder2010caltech} and FGVC-aircraft \cite{maji13fine-grained}. 
	As shown in Tab \ref{tab:cls}, our FT significantly boosts the transfer performance on iNat-18, with $1.7\%$ and $0.9\%$ improvement based on MoCo and MoCo-V2. Besides, our FT brings consistent improvement on CUB-200 and FGVC-aircraft.

	{\noindent \textbf{Object Detection}}
	Recent works \cite{wang2020dense,xie2021detco,xie2020propagate,yang2021instance, zhao2020makes} have shown that the transfer accuracy of state-of-the-arts (SOTAs) models \cite{caron2020unsupervised,chen2020simple,tian2020makes,chen2020improved,he2020momentum}  on classification and detection are inconsistent and have low correlation, denoted as ``task-bias''. One important reason is that pre-tasks of SOTA are specifically designed and optimized for classification, such as instance discrimination \cite{wu2018unsupervised,he2020momentum} and clustering \cite{caron2020unsupervised}, leading to substantial enhancement on classification but slight gain for detection. 
	Therefore we evaluate our FT on detection/instance segmentation tasks. As summarized in Tab~\ref{tab:final}, our FT can boosts the baseline model MoCo-V2 on various datasets and metrics respectively. Our FT strongly improves the transfer accuracy] on VOC \cite{everingham2010pascal_VOC} and MSCOCO \cite{lin2014microsoft_coco}. 
	Besides, our FT with MoCo-V2 can get slightly better accuracy than those contrastive models specifically designed for detection tasks, \eg, DetCo\cite{xie2021detco} and InsLoc \cite{yang2021instance}.  Moreover, our FT can get much better classification results than DetCo.  
	Notice that our FT is not aiming at the local information during pre-task design, but more invariance from feature transformation.
	These experiments indicate that our FT is less task-bias than the pre-task-based contrastive models.
	The performance boosts suggest the efficacy and robustness of our proposed FT, and enable us to learn more ``view-invariant'' and discriminative representations.

	\section{Conclusions}
	In this work, we have developed a visualization tool to visualize the score distributions of positive and negative pairs. 
	Leveraging this visualization tool, we can understand the inside of the contrastive learning process. 
	More specifically, we discover significant observations inspiring our novel Feature Transformation, including positive extrapolation such that more hard positives are created for the training. Besides, we propose the interpolation among negatives, which makes full use of negatives and provides diversified negatives.  The feature transformations enable to learn more view-invariant and discriminative representations. Experiments show that our proposed Feature Transformation can improve at least $6.0\%$ accuracy on ImageNet-100 over MoCo, and about $2.0\%$ accuracy on ImageNet-1K over the MoCoV2 baseline. Transferring to the downstream tasks successfully demonstrate our model is less task-bias. In our future work, we will explore more feature manipulation strategies with the help of our  visualization tool.

	\clearpage
	
	{\small
		\bibliographystyle{ieee_fullname}
		\bibliography{egbib}
	}

	\clearpage
	\onecolumn
	\begin{center}
		\LARGE
		\textbf{Appendix}\\\
		\\
		
	\end{center}
	
	\maketitle
	
	\appendix
	
	\tableofcontents
	
	\clearpage
	\section{Details of the Visualization Tool}
	
	\begin{figure}
		\centering
		\vspace{-1.4cm}
		\subfigure[Mean of positive pair scores.]{
			\label{fig:supp_vision_pos_score}
			\includegraphics[width=1\linewidth,height=0.45\linewidth]{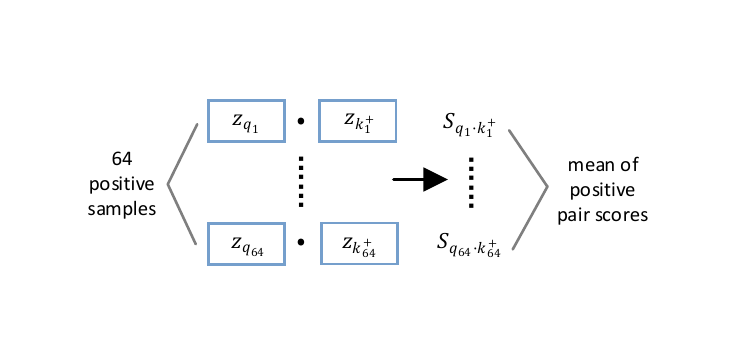}}
		\subfigure[Mean and variance of negative pair scores.]{
			\label{fig:supp_vision_neg_score}
			\includegraphics[width=1\linewidth,height=0.22\linewidth]{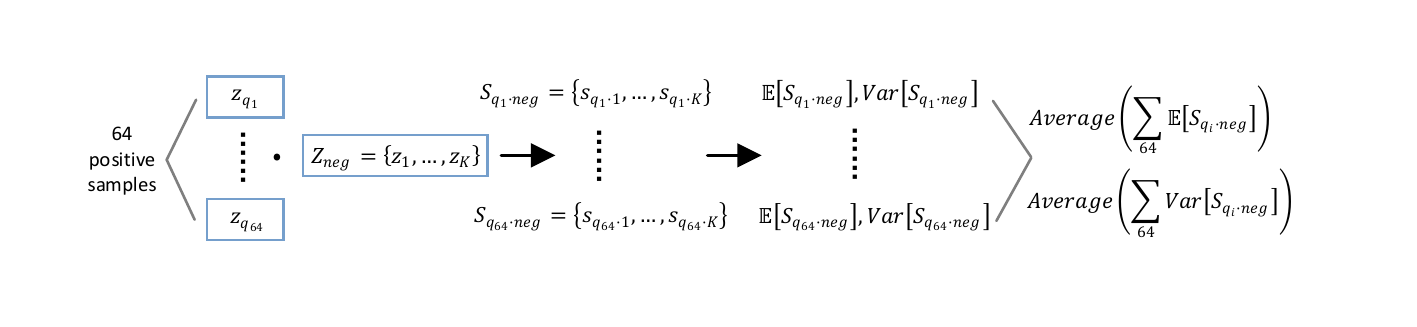}}
		
		\caption{Illustrations of pos/neg pair score visualization. (a) Mean of positive pair scores. (b) Mean and variance of negative pair scores. }
		\label{fig:supp_visual_score}
	\end{figure}
	We choose three non-trivial statistics to visualize the score distribution:
	the mean of pos/neg scores (denoted as $Mean(pos)$/$Mean(neg)$, indicating the approximate average of the pos/neg pair distance) and the variance of negative scores (denoted as $Var(neg)$, indicating the fluctuation degree of the negative samples in the memory queue).
	
	Without loss of generality, we randomly choose $64$ samples \footnote{We usually apply batch-size 256 on 4-GPU servers. Here we collect one batch $64$ on a single GPU for statistics.} in one batch to calculate the statistics data and perform visualization. 
	(1) For the positive pair score: 
	As shown in Fig \ref{fig:supp_vision_pos_score}, we denote the $z_{q_{i}}$, ($i=1, 2, 3...64$)  as the 64 query samples. And $z_{k^+_{i}}$,  are the corresponding positive features of $z_{q_{i}}$. Then we can get $64$ positive score $S_{q_i\cdot k^+_{i}}$, by inner product. Finally, we retain the mean value of these $64$ positive scores as $Mean(pos)$. 
	(2) For the negative pair scores:
	As shown in Fig \ref{fig:supp_vision_neg_score},  we denote the $Z_{neg}=\{z_{1},z_{2}, \dots, z_{K}\}$ where $K$ is the size of the memory queue \footnote{$K$ is a large number, \eg, $65536$ in MoCo \cite{he2020momentum} and the largest $8192 \times 2$ in one batch for SimCLR \cite{chen2020simple}. We use $K=65536$  in all the MoCo experiments.}.
	Each $z_{q_{i}}$ combining $Z_{neg}$ will create $K$ negative pair scores in a set, named  $S_{q_i\cdot neg}=\{s_{q_i\cdot 1},s_{q_i\cdot 2}, \dots, s_{q_i\cdot K}\}$.  
	To keep all the $64 \times K$ negative scores is challenging (about 4TB storage for the pair scores), 
	so for each $S_{q_i\cdot neg}$, we retain their mean and variance to show the distribution of $K$ negative sample scores corresponding to $z_{q_{i}}$. 
	More generally, we further average these $64$ means and variances to show the statistical characteristics of these $K$ negative samples ($Mean(neg)$ and $Var(neg)$). These statistics are recorded at each training step to track the score distribution in the training process. 
	
	Our visualization is very practical. It is offline, which almost does not affect the training speed. Instead of storing K (65536) pair scores, we save their statistical mean \& variance to represent the scores’ distribution. As a result, it only takes about 20MB storage and 5 minutes extra time for a 256 batch-size 100 epoch training. Even with larger datasets and batch size, it’s still feasible.

	\section{Experimental Details of each Part in the Paper}
	The experiments are mainly implemented using the code from InfoMin~\cite{tian2020makes}~\footnote{https://github.com/HobbitLong/PyContrast}.
	The transfer experiments on object detection and instance segmentation are implemented using Detectron2~\footnote{https://github.com/facebookresearch/detectron2}. 	We keep the fairness of the experiments, especially when compared with other methods.
	The code of our proposed methods and visualization tools will be made public.

	\subsection{Data Augmentations}
	For the experiments of combining our feature transformation module with other contrastive learning methods, we use the same image-level data augmentation strategies as the respective methods.
	Specifically, for our visualization experiments and other experiments using MoCo, we use the same data augmentation strategies with MoCo which contains \textit{Random Resized Crop}, \textit{Horizontal Flip}, \textit{ColorJitter}, and \textit{Random Gray Scale}.
	For the experiments on MoCoV2~\cite{chen2020improved} and SimCLR~\cite{chen2020simple}, the data augmentation strategies are the same which contains \textit{Random Resized Crop}, \textit{Horizontal Flip}, \textit{ColorJitter}, \textit{Random Gray Scale}, and \textit{Gaussian Blur}.

	\subsection{Implementation of Visualization experiments}
	{\noindent \textbf{For training}}
	All the visualization experiments are carried on ImageNet-100 and ResNet-18 for fast evaluation and parameters-tuning experiments.
	For the visualization experiments (including Table 1, Table 6 (2nd row), figure 1(a),3,4,5,7 in the paper and Table 11 (2nd row), 14,  figure 9,10,11 in supplementary materials), we apply a mini-batch size of 256 is used with 4-GPUs, where the number of negative examples is set to 65,536, with initial learning of 0.03. And we use $256/4=64$ samples to perform visualizations.  For the fast grid experiments, the model is trained for only 100 epochs with the learning rate multiplied by 0.1 at 60 and 80 epochs.
	We use SGD as the optimizer, the weight decay of SGD is 0.0001 and the momentum of SGD is 0.9. And for various unit-sphere projection experiments, we apply  200 epochs training to perform visualization. 
	
	{\noindent \textbf{For testing}}
	we use the linear readout protocol to evaluate the trained representation on the validation set by fixing the learned representation and train a supervised linear classifier on the representations, the single-crop top-1 accuracy on the validation set is reported. An initial learning rate of 10 and weight decay 0. 
	The classifier is trained with 100 epochs and the learning rate is multiplied by 0.1 at  60, and 80 epochs.

	\subsection{Implementation on ImageNet-100}
	{\noindent \textbf{For training}}
	we use ResNet-50 for ImageNet-100 implementations
	And momentum parameter is set to be 0.99 for our experiments. (including Table 2,3,4,5,6 (2nd row),7 in the paper and Table 11 (1st row),12,15,16 in supplementary materials). 
	A mini-batch size of 256 is used with 8-GPUs, where the number of negative examples is set to 65,536, with  initial learning of 0.03. 
	The model is trained for 200 epochs with the learning rate multiplied by 0.1 at 120 and 160 epochs.
	We use SGD as the optimizer, the weight decay of SGD is 0.0001 and the momentum of SGD is 0.9.

	{\noindent \textbf{For testing}}
	we use the linear readout protocol to evaluate the trained representation on the validation set by fixing the learned representation and train a supervised linear classifier on the representations, the single-crop top-1 accuracy on the validation set is reported.
	We use an initial learning rate of 10 and weight decay 0. 
	The classifier is trained with 60 epochs and the learning rate is multiplied by 0.1 at 30, 40, and 50 epochs following ~\cite{shen2020mix}.

	\subsection{Implementation on ImageNet-1k}
	
	{\noindent \textbf{For training}}
	The momentum update parameter $m$ for the experiments on ImageNet-1k is set to 0.999, other parameters are set to the same as the experiments on ImageNet-100. ResNet-50 is used as an encoder. (including Table 8 in the paper and Table 13 in supp material). 
	We can observe that the best result for positive extrapolation and negative interpolation is achieved when $\alpha_{in}$ and $\alpha_{ex}$ are set to 1.6 and 2.0 respectively. Thus we use this value for the other experiments. Except otherwise stated, other hyper-parameters are set to be the same with MoCo \cite{he2020momentum} and MoCoV2\cite{chen2020improved}.

	{\noindent \textbf{For testing}}
	The same linear readout protocol is used where the linear classifier is trained for 100 epochs and the initial learning rate is 30 which are multiplied by 0.1 at 60, 80epochs.

	\subsection{Implementation on Fine-grained Classification}
	In addition to object detection and instance segmentation tasks, we also provide a study of fine-grained classification.
	We choose three challenging fine-grained datasets to conduct the experiments, iNaturalist 2018 dataset, CUB-200 dataset, and FGVC-aircraft dataset.
	(1) The iNaturalist 2018 has ~437k images and 8142 classes, this dataset is commonly used for fine-grained classification and long-tailed recognition, and is used by several papers for evaluating the transfer performance of self-supervised representations~\cite{he2020momentum}.
	(2) The CUB-200 dataset contains 6033 images belong to 200 bird species and is used for fine-grained classification.
	(3) The FGVC-aircraft dataset has 10,200 images of aircraft, with 100 images for each of 102 different aircraft model variants, most of which are airplanes. 
	When transferring to these datasets, the pre-trained model is fine-tuned with 100 epochs, the learning rate is set to 5e-3 with cosine decay.
	
	\subsection{Object detection on PASCAL VOC}
	The main goal of self-supervised pre-training is to obtain representation that can be beneficial for downstream tasks. 
	We choose to use PASCAL VOC~\cite{everingham2010pascal_VOC} and COCO~\cite{lin2014microsoft_coco} as our benchmark for testing the transfer performance of the representation to object detection and instance segmentation tasks following previous works \cite{he2020momentum}.
	For PASCAL VOC dataset, we use the \texttt{trainval07+12} split for fine-tuning, and the \texttt{test2007} split for evaluating. The image scale is set to [480, 800] pixels for training and 800 for testing.
	For COCO dataset, we use the \texttt{train2017} split (118k images) for fine-tuning, the \texttt{val2017} split for evaluating. The image scale is set the same with PASCAL VOC.
	
	When transferring to detection tasks, feature normalization has been shown to be crucial during fine-tuning~\cite{he2020momentum}.
	Therefore, the pre-trained backbone is fine-tuned with Synchronized BN (SyncBN)~\cite{peng2018megdet_syncbn} and add SyncBN to the FPN layer following ~\cite{he2020momentum}.
	We use Faster R-CNN~\cite{faster} with R50-C4 architectures for object detection on the PASCAL VOC dataset.
	All layers of the model are fine-tuned with 24,000 iterations with each batch consisting of 16 images.
	The initial learning is set to 0.02 and is multiplied by 0.1 at 18,000 and 22,000 iterations.
	Other hyper-parameters are set to be the same with \cite{he2020momentum}.

	\subsection{Object detection and Instance segmentation on MSCOCO}
	We also tested the transferring abilities of the pre-trained model using the instance segmentation tasks on MS COCO dataset.
	We uses a Mask R-CNN~\cite{he2017mask} R50-FPN pipeline following ~\cite{tian2020makes}.
	The batch size is set to 16 with the learning rate as 0.02, the model is trained with 1x and 2x schedules, for 1x schedules, the model is trained for 90,000 iterations on the MS COCO datasets with the learning rate multiplied by 0.1 at 60,000 and 80,000 iterations, for the 2x schedules, we use 180,000 iterations with the learning rate multiplied by 0.1 at 120,000 and 160,000 iterations.
	The transfer results of the 2x schedule is provided in Tab~\ref{tab:supp_transfer_coco_2x_fpn}. Other hyper-parameters are set to be the same with \cite{he2020momentum}.

	\begin{table*}[h]
		\centering
		\begin{tabular}{l|c|c|c|c|c|c}\hline
			\multirow{2}{*}{Method} & \multicolumn{6}{c}{Performance}                         \\  
			& AP$^{bb}$ & AP$^{bb}_{50}$ & AP$^{bb}_{75}$  & AP$^{mk}$ & AP$^{mk}_{50}$ & AP$^{mk}_{75}$ \\ \hline\hline
			mocov1                  & 40.7      & 60.5           & 44.1            & 35.4      & 57.3           & 37.6           \\
			mocov1+ours             & 41.5      & 61.0           & 44.5            & 35.9      & 57.7           & 38.0           \\
			mocov2                  & 40.9      & 60.7           & 44.4            & 35.5      & 57.5           & 37.9           \\
			mocov2+ours             & 41.3      & 60.9           & 44.8            & 35.7      & 57.8           & 38.1           \\\hline
		\end{tabular}
		\vspace{0.1cm}
		\caption{COCO object detection and instance segmentation based on Mask-RCNN-FPN with 2x learning rate schedule. Our results are reported using the average of 3 runs.}
		\label{tab:supp_transfer_coco_2x_fpn}
	\end{table*}

	\section{Details of the Gradient Landscape}
	
	\begin{figure}
		\centering
		\subfigure[$m=0.99$]{
			\label{fig:supp_t007_m099_L2grad}
			\includegraphics[width=0.3\linewidth,height=0.31\linewidth]{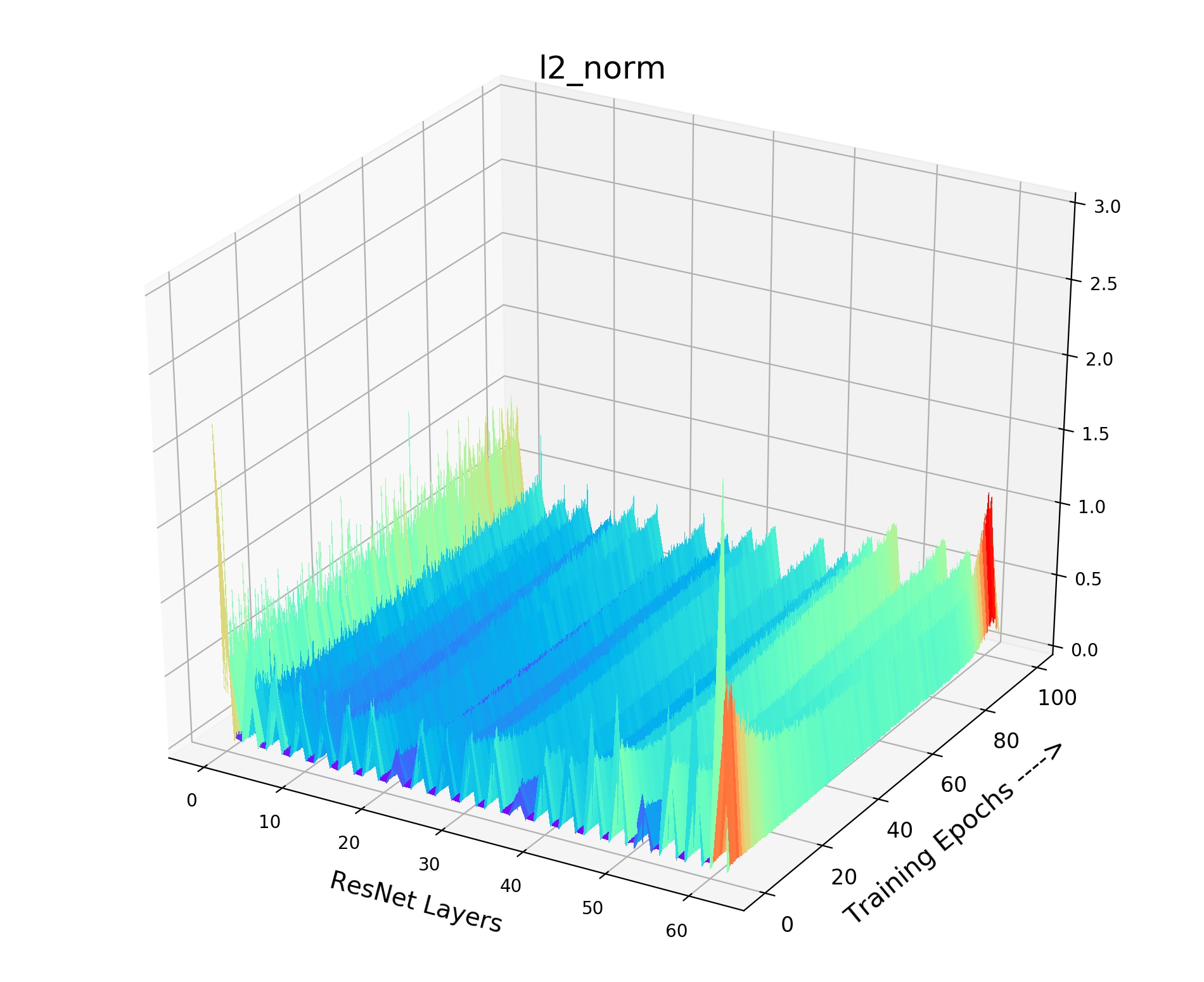}}
		\subfigure[$m=0.6$]{
			\label{fig:supp_t007_m06_L2grad}
			\includegraphics[width=0.32\linewidth,height=0.325\linewidth]{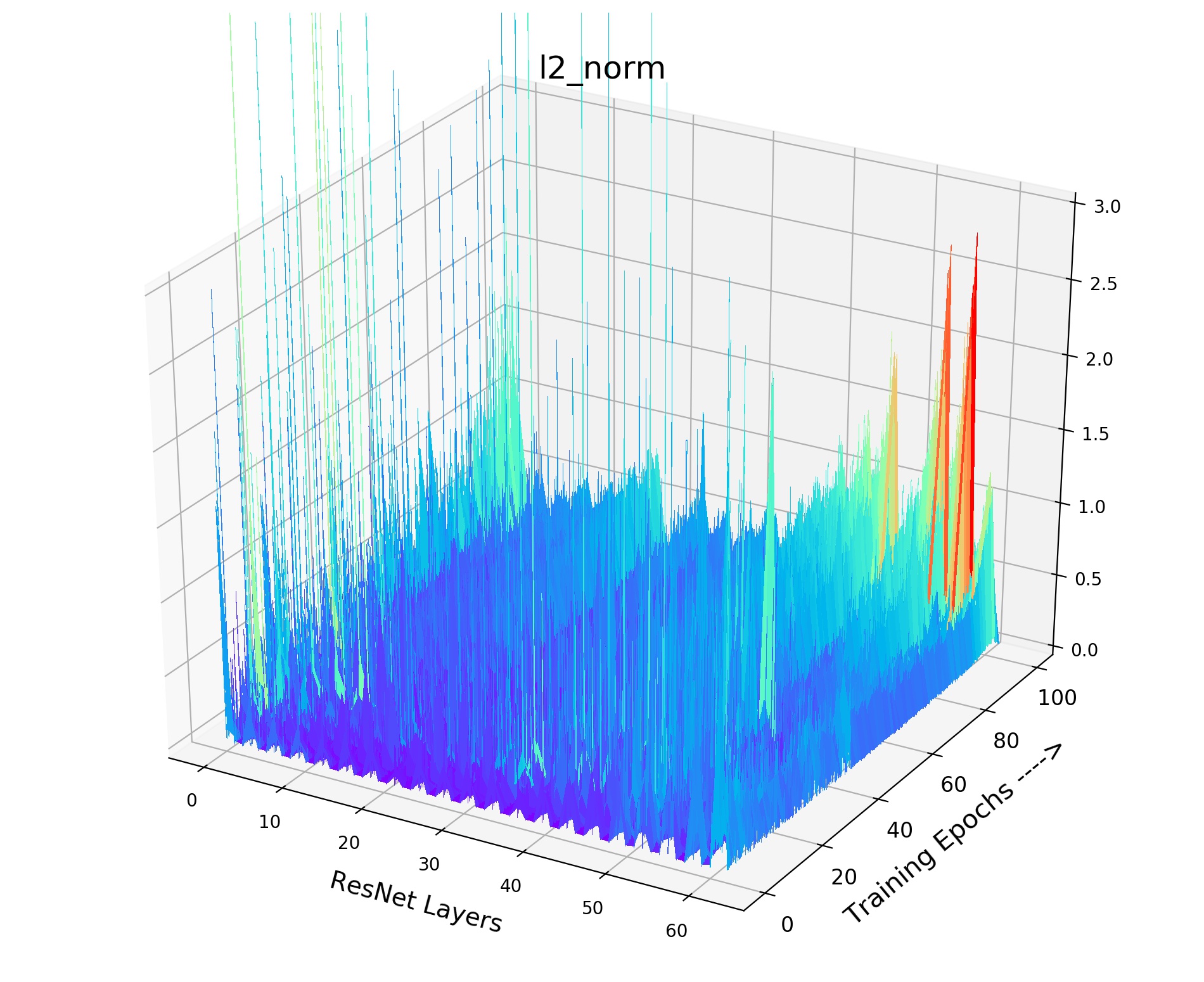}}
		\subfigure[$m=0.5$]{
			\label{fig:supp_t007_m05_L2grad}
			\includegraphics[width=0.32\linewidth,height=0.325\linewidth]{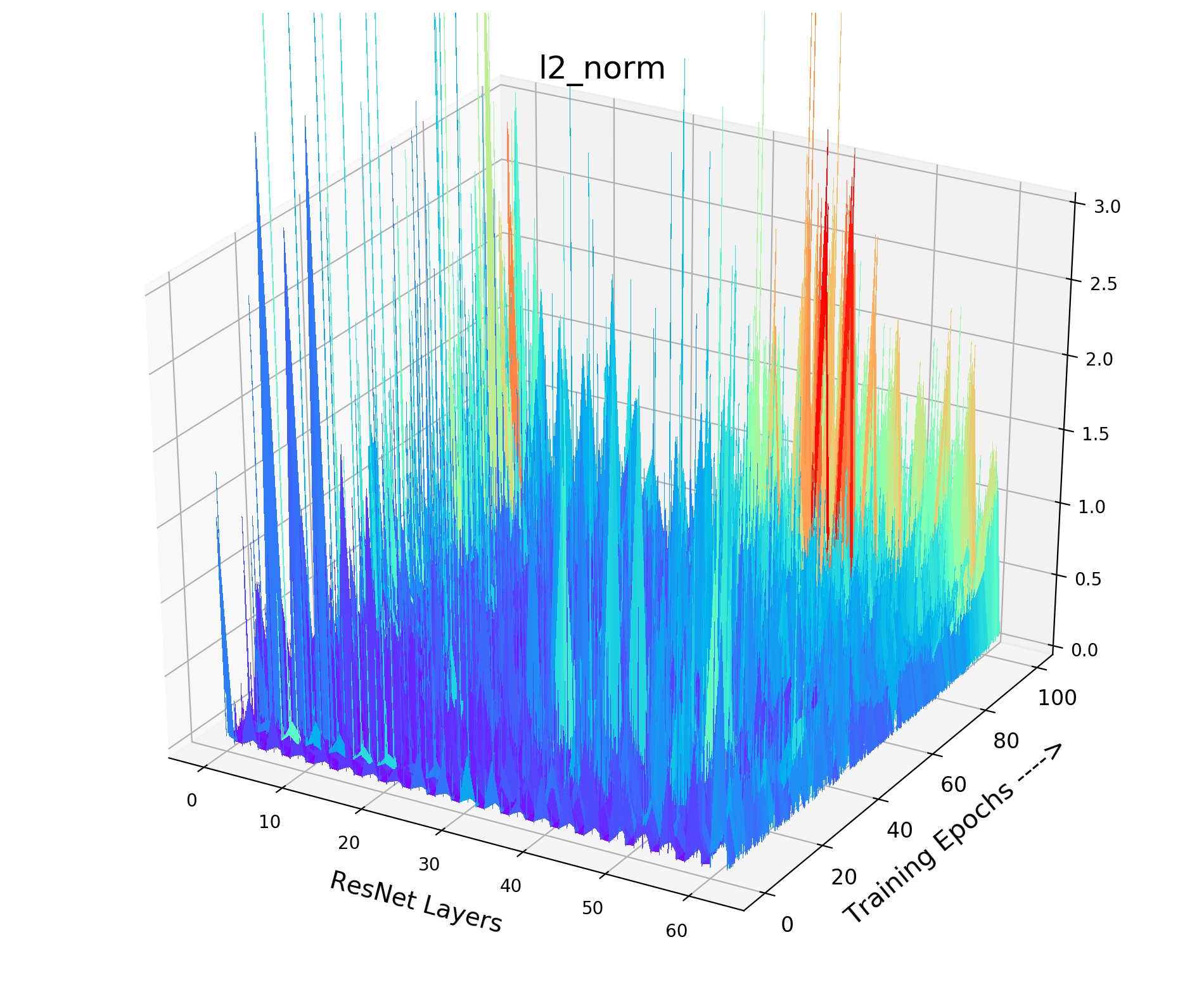}}
		\caption{Gradient ($\ell_2$ norm) landscape of various $m$. We provide $\ell_2$ norm for  each layer of the encoder (ResNet-18)  with the training process. Across Y axis indicating the training epochs. Across X axis representing the ResNet layers, it shows the gradients of all layers including the BatchNorm layer whose gradient is small. But the gradient of Convolution layer is  large, thus it seems to be spiky across X axis. And Z axis means the value of $\ell_2$ norm. 
			The spiky gradient on X axis doesn't influence the training, while the smooth gradient on Y axis matters.	
			We can see that small $m=0.6$ and $0.5$ brings drastic volatility with the training process. The corresponding loss value and the gradient will fluctuate violently, resulting in bad convergence.} 
		\label{fig:supp_t007_grad}
	\end{figure}
	
	We provide the details of our gradient landscape Figure 4 of various $m$ in the paper. 
	As shown in Fig \ref{fig:supp_t007_grad}, we provide $\ell_2$ norm for  each layer of the encoder (ResNet-18)  with the training process. X axis indicates the layers of the encoder, while Y axis indicates the 100 training epochs. And Z axis means the value of $\ell_2$ norm.
	We choose the $\ell_2$ norm of this layer (total gradient $\ell_2$ norm of this layer) because the $\ell_2$ norm of gradient is very obvious to show the smoothness of gradient landscape. We can see that small $m=0.6$ and $0.5$ brings drastic volatility with the training process. The corresponding loss value and the gradient will fluctuate violently, resulting in bad convergence. As shown in Fig \ref{fig:supp_t007_grad}, the smooth and stable gradient landscape of $m=0.99$ (Fig \ref{fig:supp_t007_m099_L2grad}) becomes sharp and messy with the decrease of $m$ (Fig \ref{fig:supp_t007_m06_L2grad} for $m=0.6$ and Fig \ref{fig:supp_t007_m05_L2grad} for $m=0.5$). Therefore, to learn a better pre-trained model, we need to prepare negative pairs that can maintain the stability and smoothness of score distribution and gradient for the training process. 
	It seems that the gradient landscape looks spiky:  
	1) Across Y axis indicating the training epochs.
	2) Across X axis representing the ResNet layers, it shows the gradients of all layers including the BatchNorm layer whose gradient is small. But the gradient of Convolution layer is  large, thus it seems to be spiky across X axis. 
	The spiky gradient on X axis doesn't influence the training, while the smooth gradient on Y axis matters.

	\begin{figure}
		\centering
		\subfigure[Mean of neg scores in early stage]{
			\label{fig:supp_whentoadd_negmean_initial}
			\includegraphics[width=0.3\linewidth,height=0.3\linewidth]{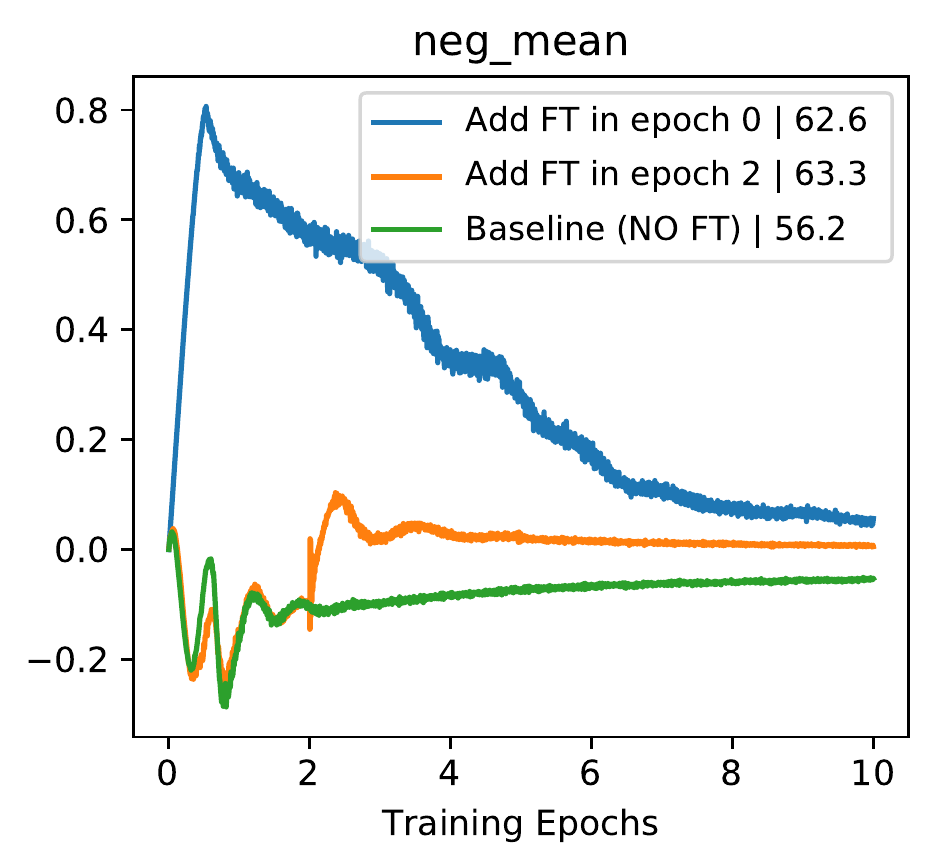}}
		\subfigure[Mean of pos scores in early stage]{
			\label{fig:supp_whentoadd_posmean_initial}
			\includegraphics[width=0.3\linewidth,height=0.3\linewidth]{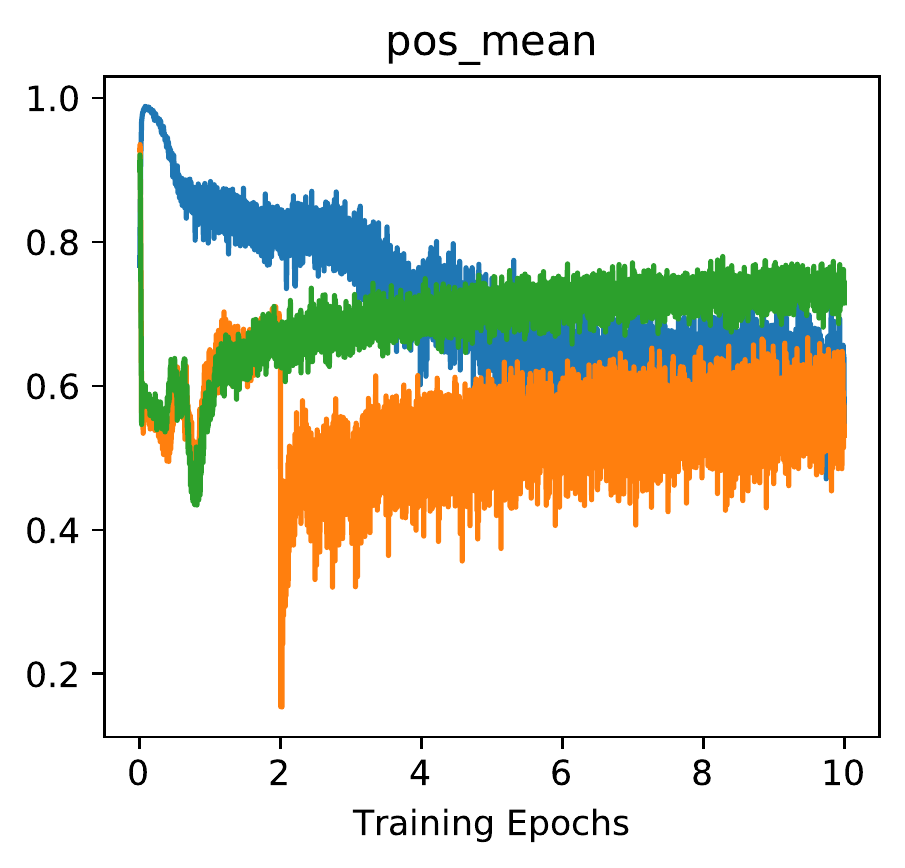}}
		\subfigure[Mean of neg scores]{
			\label{fig:supp_whentoadd_negmean}
			\includegraphics[width=0.3\linewidth,height=0.3\linewidth]{whentoadd/when_to_add_negmean.pdf}}
		\subfigure[Mean of pos scores]{
			\label{fig:supp_whentoadd_posmean}
			\includegraphics[width=0.3\linewidth,height=0.3\linewidth]{whentoadd/when_to_add_posmean.pdf}}
		\subfigure[Baseline MoCo landscape]{
			\label{fig:supp_whentoadd_baseline}
			\includegraphics[width=0.3\linewidth,height=0.3\linewidth]{whentoadd/whentoadd_baseline_l2norm.jpg}}
		\subfigure[Adding FT in 0th epoch]{
			\label{fig:supp_whentoadd_0ep}
			\includegraphics[width=0.3\linewidth,height=0.3\linewidth]{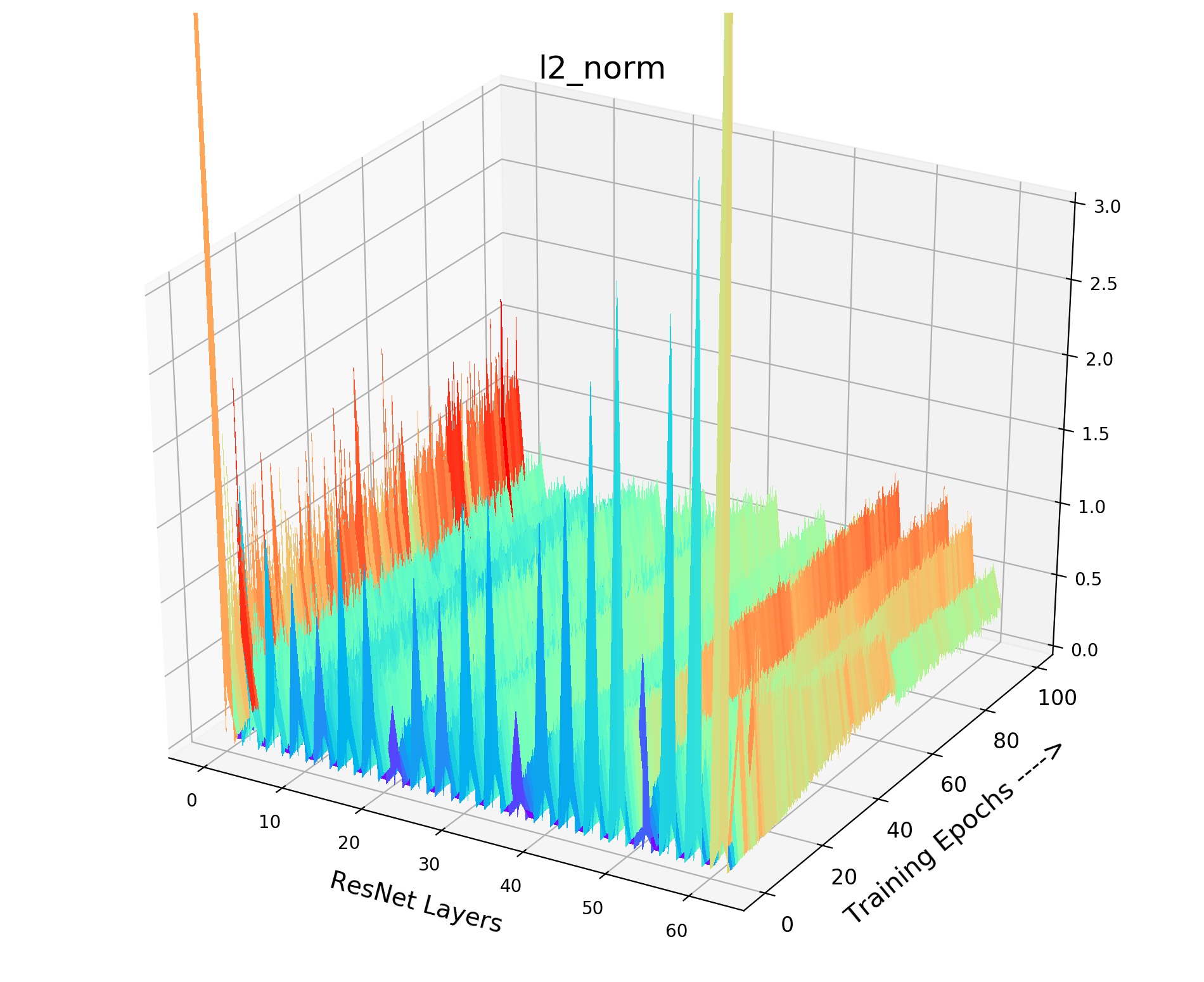}}
		\subfigure[Adding FT in 2nd epoch]{
			\label{fig:supp_whentoadd_2ep}
			\includegraphics[width=0.3\linewidth,height=0.3\linewidth]{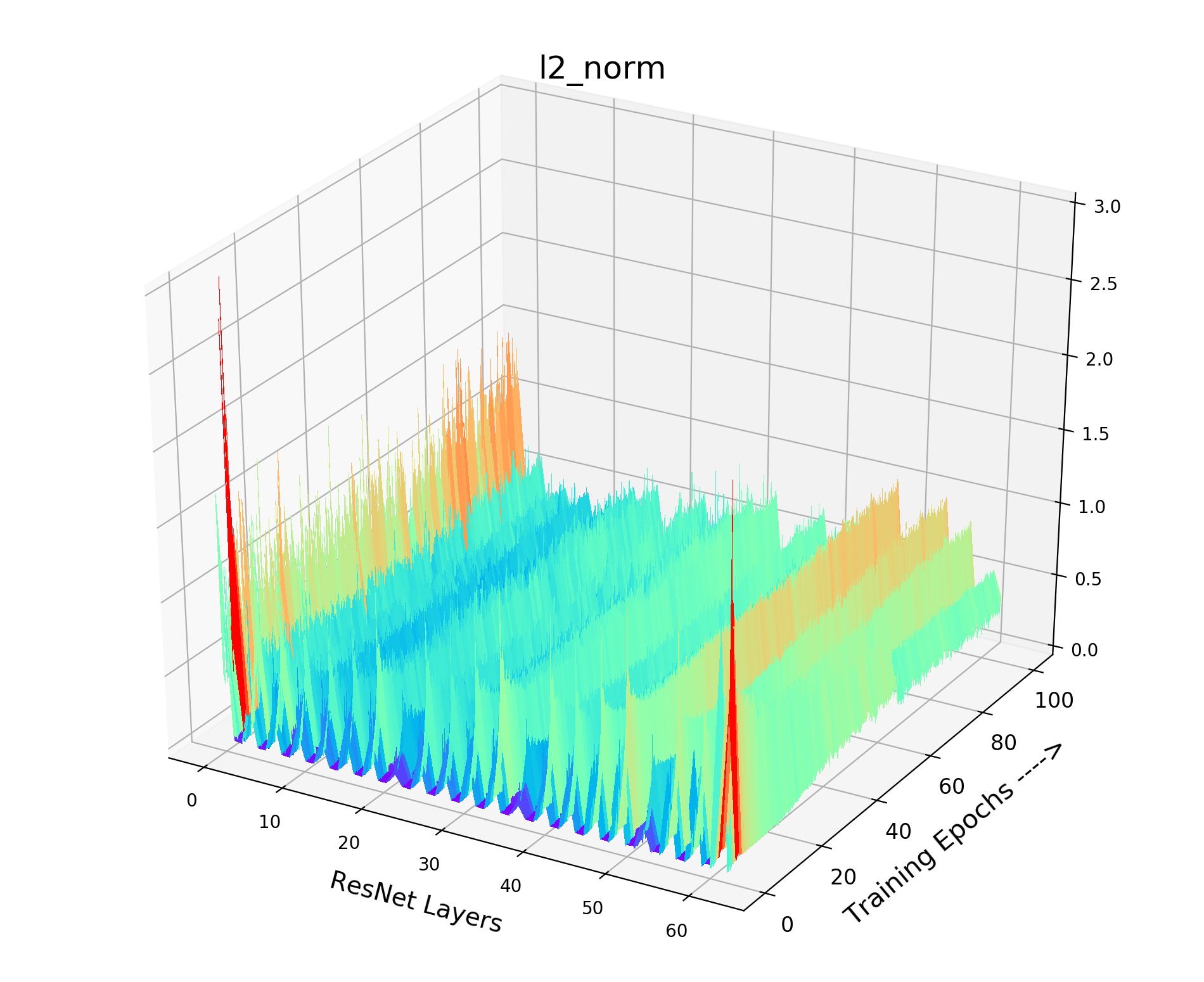}}
		\subfigure[Adding FT in 30th epoch]{
			\label{fig:supp_whentoadd_30ep}
			\includegraphics[width=0.3\linewidth,height=0.3\linewidth]{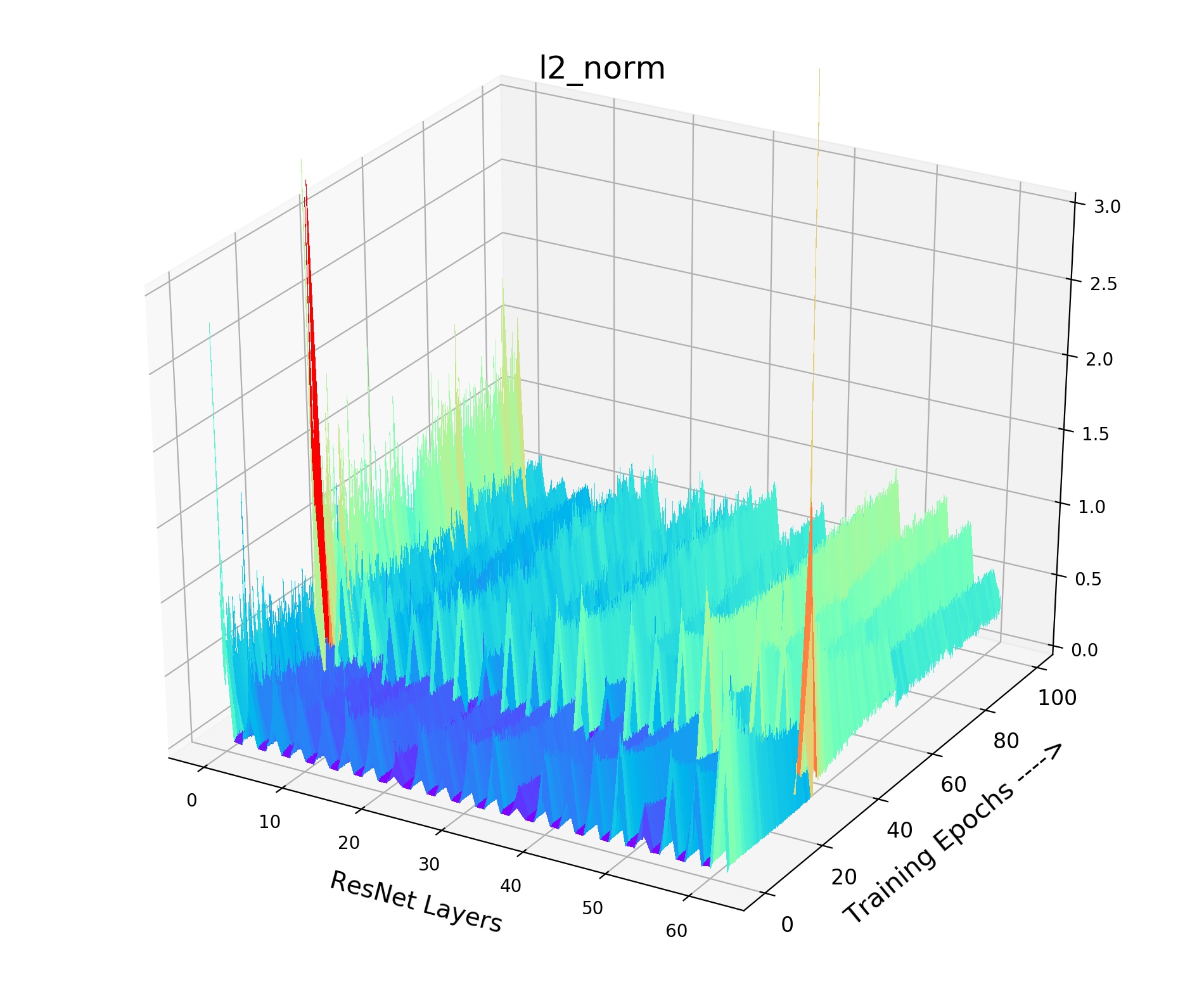}}
		\subfigure[Adding FT in 50th epoch]{
			\label{fig:supp_whentoadd_50ep}
			\includegraphics[width=0.3\linewidth,height=0.3\linewidth]{whentoadd/whentoadd_50ep_l2norm.jpg}}
		%\vspace{-0.1cm}
		\caption{Visualization of when to add FT, including score distribution and Gradient ($\ell_2$ norm) landscape.}
		\label{fig:supp_whentoadd}
	\end{figure}
	
	\section{Discussion of when to Add the Feature Transformation}
	
	\subsection{Effectiveness of our FT}
	We present the efficacy of FT by analysis of starting FT in various training stages. As shown in Tab \ref{tab:supp_in100_whentoadd}, starting FT (pos extrapolation + neg interpolation) from various epoch can boost the accuracy of baseline, and starting from earlier can improve more ($7.1\%$/$5.8\%$ boosts with Res-18/Res-50). It is worthy to note that even adding FT in the 80th epoch can bring $3\%$ improvement compared with the MoCo baseline (No FT in training).
	With the visualizations of score distribution Fig \ref{fig:supp_whentoadd}, we can see that our FT not only brings hard positives (lowering pos scores in Fig \ref{fig:supp_whentoadd_posmean}) and hard negatives (rising neg scores in Fig \ref{fig:supp_whentoadd_negmean}) simultaneously when the combined FT is inserted in various stages. The combination of positive extrapolation and negative interpolation can help rise the neg scores in the training process. 
	Besides, with the comparison of the Gradient ($\ell_2$ norm) landscape, we can observe that our FT brings a greater gradient for the training (Adding FT in the 30th epoch Fig \ref{fig:supp_whentoadd_30ep} and 50th epoch Fig \ref{fig:supp_whentoadd_50ep}), which makes the model escape from the local minima and avoid over-fitting. These analyses indicate our FT is a plug-and-play method and brings persistent view-invariance and discrimination for the training of contrastive models.

	\subsection{FT in the Early Training Stage}
	Due to the memory queue is initialized by random vectors at the start of training, the positive score and negative score have confusion, as shown in the visualizations in the early training stage (Fig \ref{fig:supp_whentoadd_negmean_initial} and Fig \ref{fig:supp_whentoadd_posmean_initial}). We provide the visualizations in the first 10 epoch to see the score distribution: 
	(1) Adding FT from the 0th epoch will bring negative pairs whose score is very high (blue line in Fig \ref{fig:supp_whentoadd_negmean_initial}, $0.8$ negative score, which is too large for negative pairs), indicating the feature transformations for the random vectors will hurt the pair score distribution. From the perspective of gradient landscape in Fig \ref{fig:supp_whentoadd_0ep}, the initial gradient brought by FT is too sharp and not smooth for training compared with the baseline MoCo in Fig \ref{fig:supp_whentoadd_baseline}. 
	(2) Adding FT from the 2nd epoch (In the 2nd epoch, the memory queue is filled by the semantic features from training data rather than the random vectors) will relieve solve too high negative scores (orange line in Fig \ref{fig:supp_whentoadd_negmean_initial}, normal negative score) and meanwhile lower the positive score from easy positive to hard one (orange line in Fig \ref{fig:supp_whentoadd_posmean_initial}, decreasing the positive score). The gradient (Fig \ref{fig:supp_whentoadd_2ep}) seems more smooth and stable compare with starting FT from 0th epoch (Fig \ref{fig:supp_whentoadd_0ep}). More importantly, in Tab \ref{tab:supp_in100_whentoadd}, starting from the 2nd epoch ($63.3\%$) can achieve slightly better accuracy than that at the beginning ($62.6\%$).
	However, in the final experiments of imagenet-1K, we still use the strategy of starting FT from the 0th epoch. Because there seems no obvious 
	performance difference in the ResNet-50 backbone in Tab \ref{tab:supp_in100_whentoadd}. Future work will focus more on this issue.
	
	\begin{table}
		\vspace{-.5em}
		\centering
		\small
		\begin{tabular}{c|cccccc}
			\toprule
			FT begin epoch & 0  &  2    & 30   & 50   & 80   & -   \\
			\midrule
			Res18 acc (\%) & 62.6 & \textbf{63.3}  & 62.9 & 61.8 & 59.2 & 56.2    \\
			Res50 acc (\%) & \textbf{76.9} & 76.4  & 75.9 & 74.0 & 72.2 & 71.1    \\
			\bottomrule
		\end{tabular}
		\vspace{0.1cm}
		\caption{\label{tab:supp_in100_whentoadd} When to add feature transformation.  We employ Res-18 (total 100 epochs) and Res-50 (total 200 epochs) on IN-100 for the results. '-' indicates MoCo baseline without using any FT.}
	\end{table}
	
	\section{Details of Comparison to other Methods}
	%% infomin, SwAV, SimSiam
	In this section, we discuss the details of how to apply our feature transformation to other self-supervised methods.
	We evaluate the performance of feature transformation on three representative methods, namely InfoMin, SwAV, and SimSiam.
	
	For feature transformation on InfoMin, we perform both positive extrapolation and negative interpolation. Note that we perform the feature transformation on both branches of the InfoMin method, i.e. the original branch and the JigSaw branch.
	For feature transformation on SwAV, we only transform the two features of the input image by positive extrapolation, the rest of the SwAV pipeline is left unchanged.
	For SimSiam, as the method only uses positive pairs for training, so we only apply the positive extrapolation as the feature transformation.
	All the other hyperparameters are set to be the same as the original paper of each self-supervised method.

	\subsection{Additional experiments of our proposed feature transformation methods on SimCLR}
	To demonstrate the effectiveness of our feature transformation methods (Negative feature interpolation and Positive feature extrapolation), we also provide the experimental results on ImageNet-100~\cite{tian2019contrastive} of applying our method on another classic contrastive learning model, SimCLR~\cite{chen2020simple}.
	Instead of using two encoders for encoding $q$ and $k$ like in MoCo~\cite{he2020momentum}, SimCLR directly uses a single network to encode the two views and contrast them against other negative examples.
	Because both MoCo and SimCLR are contrastive-based methods, the negative interpolation and positive extrapolation strategies can also be applied to SimCLR.
	We show the results of combining negative interpolation and positive extrapolation in Tab~\ref{tab:supp_in100_simclr}.
	
	\subsection{Apply Positive Extrapolation on Non-contrastive Models on IN-1K}
	Here we complement the results of applying positive extrapolation on non-contrastive models \cite{grill2020bootstrap, chen2020simple}. The models are pre-trained for 100 epochs on IN-1K with the same data augmentation setting of the original paper. As shown in Table \ref{tab:supp_compared_table}, we provide the IN-1k results (100ep) of BYOL/BYOL+posFT (66.5\% -$\textgreater$ 67.2\%)  and SimSiam/SimSiam+posFT (68.1\% -$\textgreater$ 68.7\%) indicating  pos extrapolation alone can help BYOL and SimSiam.
	Notice that we didn't perform the parameter experiments (not the optimal extrapolation parameter $\alpha_{ex}$), so the improvement is slight.

	\begin{table}[]
		\centering
		\begin{tabular}{@{}lll@{}}
			\toprule
			method & arch & acc   \\
			\midrule
			SimCLR   & r50  & 74.32 \\
			SimCLR+pos extrapolation       &  r50    &    75.80   \\
			SimCLR+neg interpolation       &  r50    &    76.71   \\
			SimCLR+both                    &  r50    &    78.25   \\
			SimCLR+both$_{dimension}$      &  r50    &    \textbf{78.81}   \\
			\bottomrule
		\end{tabular}
		\vspace{0.1cm}
		\caption{\label{tab:supp_in100_simclr} Performance comparison of our proposed two feature transformation module on imagenet-100 with SIMCLR, the model are trained for 200 epochs. 
			%the line with both$_{norm}$ is normalizing the mixed feature to the unit sphere, which show no improvements. 
			the last line with both$_{dimension}$ is mixing the feature (both pos/neg) using the dimension-level mixing, which shows improvements over the feature-level mixing.}
		\vspace{-0.2cm}
	\end{table}

	\begin{table}[]
		\centering
		\setlength\tabcolsep{1pt}
		\begin{tabular}{@{}lcc@{}}
			\toprule
			Method &   ~SimSiam~ & ~BYOL~ \\
			\midrule
			baseline$^*$    & 68.1   & 66.5 \\
			+pos extrapolation      & 68.7  & 67.2  \\
			\bottomrule
		\end{tabular}
		\vspace{0.1cm}
		\caption{\label{tab:supp_compared_table} Comparision studies of proposed methods with non-contrastive methods. The models are pre-trained for 100 epochs with Res50 on IN-1K. 
			$^*$ indicates reproduced baseline results.}
	\end{table}
	
	\section{Discussion of the feature normalization for FT}
	Here we provide additional visualization and analysis on the regular Feature Transformation (feature normalization, $\ell_2$ normalization) due to its significant constriction (unit-sphere projection) and Whether to add $\ell_2$ Normalization after our proposed FT. 
	
	\begin{figure}
		\begin{center}
			\includegraphics[width=1.0\linewidth]{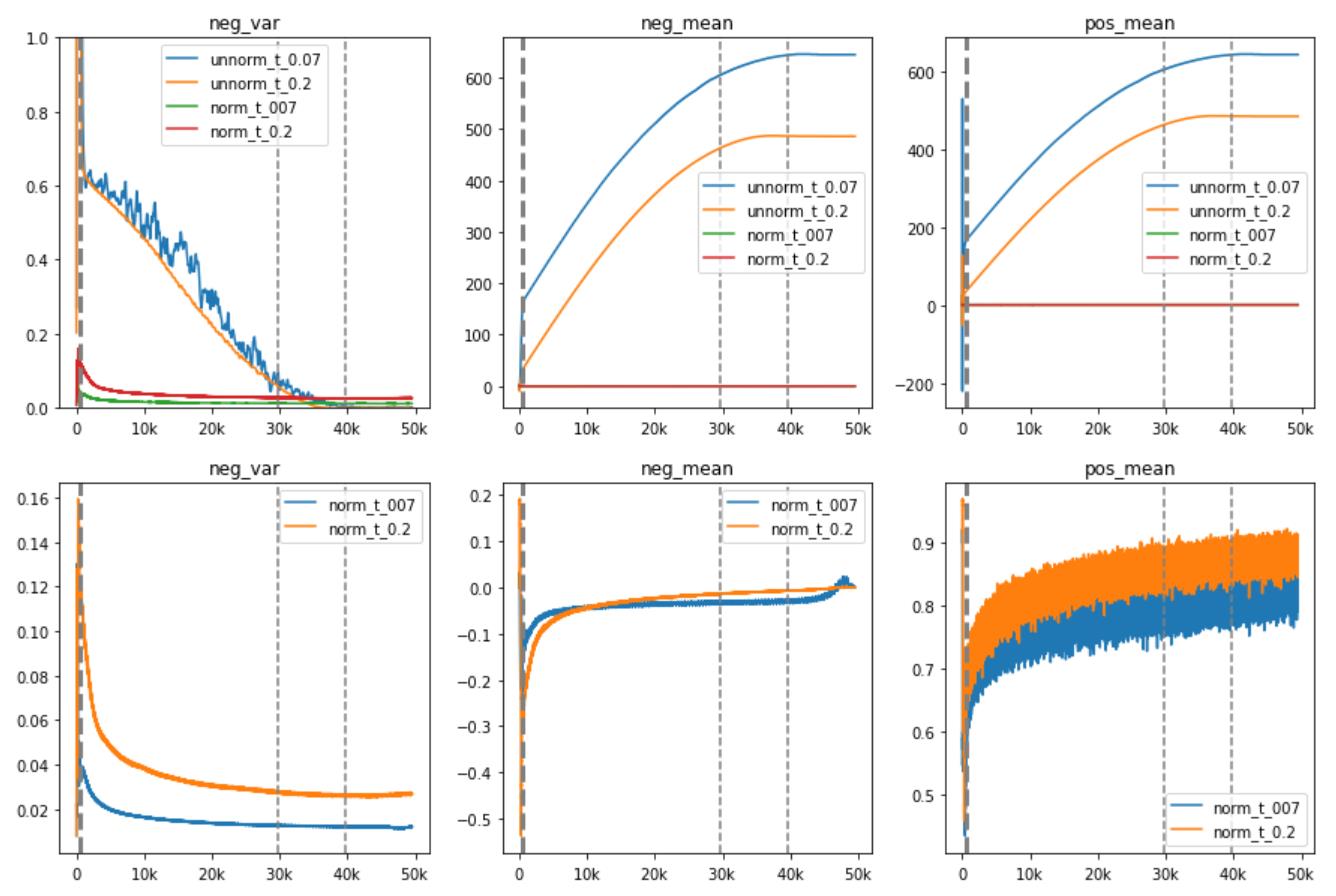}
		\end{center}
		\caption{Pos/neg pair score distribution of unit sphere projection on ImageNet-100}
		\label{fig:supp_unit}
	\end{figure}

	\subsection{Importance of Unit-sphere Projection}
	Unit-sphere projection ($\ell_2$ norm) constricts the feature vector length from unbounded to $1$, in the meanwhile retains the vector direction. Thus the pair scores $S_{q\cdot k}$ can be limited to $[-1,1]$. 	
	Recent paper \cite{khosla2020supervised} concludes that unit sphere projection plays a key role in ensuring the large gradients of hard positives and negatives from the loss gradient properties. 
	However, without the unit sphere projection, the feature vector length lost the constriction to $[-1,1]$, and the too-large score distribution leads to bad contrastive learning and poor transfer performance ($65.04\%$ \emph{v.s} $model collapse$ ). 
	As shown in our empirical study (Fig \ref{fig:supp_unit} and Table \ref{tab:supp_ft_unitsphere}) of this significant FT, the mean of positive pair score is similar to the mean of negative pair score when we removed unit-sphere projection, which will lead to an awful contrastive learning process: confusing the pos/neg pairs and bad gradient landscape brought by too large score distribution.
	Meanwhile, with the constrictions of unit-sphere projection, the mean of pos/neg pair scores are as expected: $ neg \in [-0.2,0]$ and $pos \in [0.6,0.9]$, which can be discriminated by the log-softmax loss function. This limited small score distribution benefits the later contrastive learning and brought a stable training process. Finally, the variance of the negative pair score shows that model with unit-sphere projection will provide less volatile negative pairs, which is better for contrastive learning.
	
	\begin{table}[]
		\centering
		\begin{tabular}{@{}llll@{}}
			\toprule
			Method & $\tau$ & lr & $ Acc\%$   \\
			\midrule
			MoCo w/ unit sphere proj   & $0.07$ & $0.03$  & $65.04$ \\
			MoCo  w/ unit sphere proj     & $0.2$  &  $0.03$    &    $63.06$   \\
			MoCo w/o unit sphere proj      & $0.07$  &  $0.03/10$    &    $collapse$   \\
			MoCo w/o unit sphere proj       & $0.2$  &   $0.03/10$    &    $collapse$    \\
			\bottomrule
		\end{tabular}
		\vspace{0.1cm}
		\caption{\label{tab:supp_ft_unitsphere}  The  experiments for unit sphere projection on ImageNet-100}
	\end{table}
	
	\begin{table}[]
		\centering
		\setlength\tabcolsep{1pt}
		\begin{tabular}{@{}lc@{}}
			\toprule
			Method (MoCov1) &  Acc\%  \\
			\midrule
			baseline$^*$    & 71.10   \\
			+pos extrapolation       &  72.80    \\
			+pos extrapolation$_{norm}$        &  72.45    \\
			+neg interpolation       &  74.64    \\
			+neg interpolation$_{norm}$      &  74.82    \\
			+both                    &  76.87   \\
			+both$_{norm}$           &  76.68   \\
			\midrule
			+neg extrapolation       &  71.84    \\
			+neg extrapolation$_{norm}$      &  71.95    \\
			\bottomrule
		\end{tabular}
		\vspace{0.1cm}
		\caption{\label{tab:supp_in100_norm} Ablation studies of proposed methods on various contrastive models. The model are pre-trained for 200 epochs with Res50 on IN-100. The line with ${norm}$ is re-normalizing the transformed feature to the unit sphere, which show no improvements. 
			$^*$ indicates reproduced baseline results.}
		\vspace{-0.2cm}
	\end{table}
	
	\subsection{Whether to add $\ell_2$ Normalization after FT}
	In this section, we provide empirical studies about whether re-perform the  $\ell_2$ norm for the transformed features after FT.
	As shown in Tab\ref{tab:supp_in100_norm},  the performance difference is negligible  for the model with/without re-performing the $\ell_2$ normalization, 
	($74.64\%$ \emph{v.s} post-norm $74.82\%$ for negative interpolation, $72.80\%$ \emph{v.s} post-norm $72.45\%$ for positive extrapolation, $76.87\%$ \emph{v.s} post-norm $76.68\%$ for combined FT). 
	So we conclude that the transformed features are not necessarily on the unit sphere (\ie has a norm of 1) due to the negligible performance difference. And in the final experiments of imagenet-1K, we do not re-perform the  $\ell_2$ norm after feature transformations.  \textbf{However, we strongly recommend to re-perform $\ell_2$ norm for the transformed features on all the datasets, for the sake of contrasting all the scores on the unit-sphere.}

	\section{Discussion of the Negative Feature Transformation}
	
	In this section, we provide more discussions about the feature manipulation of the negative examples. 
	We have discussed negative interpolation to fully utilize negative features and increase the diversity of the memory queue. Here we provide the situation about negative extrapolation in memory queue and creating hard negatives.

	\begin{table}[]
		\centering
		\setlength\tabcolsep{1pt}
		\begin{tabular}{@{}lcc@{}}
			\toprule
			Method (MoCov1) &\quad Beta parameter    &   \quad Acc\%  \\
			\midrule
			baseline$^*$   & \quad- & \quad71.10  \\
			+neg interpolation       &  \quad$Beta(1.6, 1.6)$   & \quad74.64  \\
			+neg extrapolation       & \quad$Beta(2.0, 2.0)$    & \quad71.84  \\
			+hard negative           & \quad$Beta(2.0, 1.0)$  &   \quad73.45\\
			+hard negative           & \quad$Beta(5.0, 2.0)$  &  \quad74.32 \\
			\bottomrule
		\end{tabular}
		\vspace{0.1cm}
		\caption{\label{tab:supp_in100_negative} Ablation studies of proposed methods on various contrastive models. The model are pre-trained for 200 epochs with Res50 on IN-100. The line with ${norm}$ is normalizing the transformed feature to the unit sphere, which show no improvements. 
			$^*$ indicates reproduced baseline results.}
		\vspace{-0.2cm}
	\end{table}
	
	\subsection{Negative Extrapolation in Memory Queue}
	We have explored the negative interpolation to fully utilize negative features and increase the diversity of the memory queue. Then how about the negative extrapolation in the memory queue? Will the extrapolated negatives still be effective to increase the diversity of the memory queue and the performance?
	
	Specifically, we denote the negative memory queue of MoCo as $Z_{neg}=\{z_{1},z_{2}, \dots, z_{K}\}$ where $K$ is the size of the memory queue, and $Z_{perm}$ as the random permutation of $Z_{neg}$.
	We propose to use a simple extrapolation between two memory queue to create a new queue $\hat{Z}_{neg}^{ex}=\{\hat{z}_{1}^{ex}, \hat{z}_{2}^{ex}, \dots, \hat{z}_{K}^{ex}\}$:
	\begin{equation}
	\hat{Z}_{neg}^{ex} = \lambda_{ex}\cdot Z_{neg} + (1 - \lambda_{ex})\cdot Z_{perm}
	\end{equation}
	where $\lambda_{ex} \sim Beta(\alpha_{ex}, \alpha_{ex}) + 1$ is in the range of $(1,2)$. The transformed memory queue $\hat{Z}_{neg}$ provides fresh extrapolated negatives for contrastive loss iteration by iteration.
	As shown in Tab \ref{tab:supp_in100_negative}, the negative extrapolation brings slight improvement over baseline ($71.84\%$ \emph{v.s.} $71.10$, $0.74\%$ improved), while negative interpolation significantly improves to $74.64\%$.
	Both the negative interpolation and extrapolation can increase the diversity of the memory queue, but why extrapolation cannot boost the performance?
	We conjecture that  the original queue $Z_{neg}$ provides discrete distribution of negative samples but our method can fill in the incomplete sample points of the distribution by random interpolation, leading to a more discriminative model.
	But the extrapolated sample points may not stay in the previous manifold/distribution. Future work will focus more on this discussion.

	\subsection{Creating Hard Negatives}
\begin{figure}
		\begin{center}
			\includegraphics[width=0.5\linewidth]{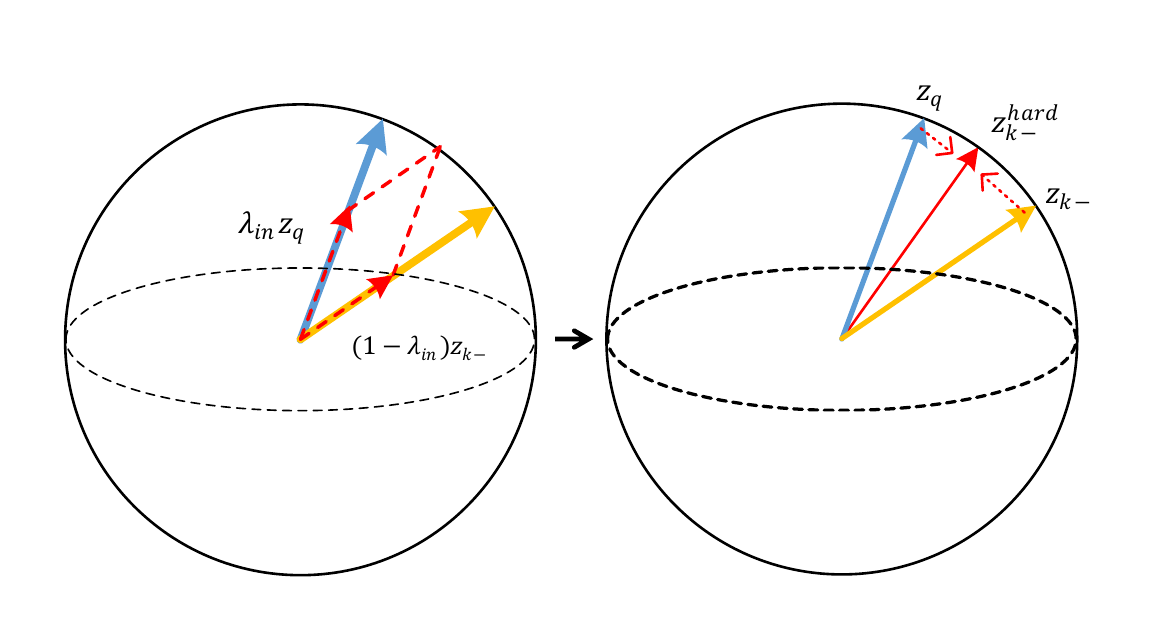}
		\end{center}
		\vspace{0.1cm}
		\caption{The process of creating hard negatives. The distance between the pos/neg feature vector is lowered, changing easy negatives to hard negatives, which is better for contrastive learning.}
		\label{fig:supp_hard_negative}
	\end{figure}

	The negative interpolation and extrapolation are both performed in the memory queue to increase the diversity. Another feature transformation for negative features is to increase the hardness during training, like the way of positive extrapolation. Our goal is to increasing the easy negative pair scores (similarity) to create hard negative pairs during training could be beneficial for the final transfer performance.
	Specifically, we use interpolation between ${z}_q$ and all the negatives in the memory queue $Z_{neg}=\{z_{1},z_{2}, \dots, z_{K}\}$ to create a hard negative queue $Z_{neg}^{hard}=\{z_{1}^{hard},z_{2}^{hard}, \dots, z_{K}^{hard}\}$. 
	\begin{equation}
	{Z}_{neg}^{hard} = \lambda_{in}\cdot {z}_q + (1 - \lambda_{in})\cdot Z_{neg}
	\end{equation}
	This equation indicates that each negative sample in the memory queue $Z_{neg}$  will be interpolated with ${z}_q$ to create hard negative queue $Z_{neg}^{hard}$. And $\lambda_{in} \sim Beta(\alpha_{in}, \alpha_{in})$ is in the range of $(0,1)$. By this transformation,  we can guarantee that the transformed neg score ${S}_{q\cdot k^{-}}^{hard}$ is larger than the original pos score $S_{q\cdot k^{-}}$, namely $z_qz_{k^{-}}^{hard} \geq z_qz_{k^{-}}$, which means we create a hard negative queue.
	Intuitively, it can be seemed like a simple approach to draw $z_q$ and $z_{k^{-}}$ closer in feature space. After interpolation, the distance between the pos/neg feature vector is lowered. Therefore this interpolation can serve as a feature transformation to create hard negatives from easy ones. As shown in Fig \ref{fig:supp_hard_negative}, it brings a minor direction change for positive/negative vectors.	As shown in Tab \ref{tab:supp_in100_negative}, our hard negatives can bring consistent boosts over the baseline ($74.32\%$ \emph{v.s.} ($71.10\%$, $3.22\%$ improved), indicating that this hard negative is effective for the contrastive learning. Future work will focus more on this topic.
	However, we choose the negative interpolation rather than the hard negative strategy in the final experiments of IN-1K. Because the computation of hard negative strategy is too large (Each  $z_q$ needs a new hard negative queue, so it takes time for one large batch to produce hard negative queue.).
	
\end{document}